\documentclass[letterpaper, 10 pt, conference]{ieeeconf}  % Comment this line out if you need a4paper

\IEEEoverridecommandlockouts                              % This command is only needed if 
\usepackage{graphicx} % for pdf, bitmapped graphics files
\usepackage{amsmath} % assumes amsmath package installed
\usepackage{amssymb}  % assumes amsmath package installed
\usepackage{subcaption}
\usepackage{siunitx}
\usepackage{multirow}
\usepackage{cite}
\usepackage{bm}
\usepackage{mathtools}
\usepackage{pifont}
\usepackage[dvipsnames]{xcolor}
\usepackage[font=footnotesize]{caption}
\usepackage[abbreviations]{glossaries-extra}
\usepackage{booktabs}
\usepackage[colorlinks = true,
            linkcolor = blue,
            urlcolor  = blue,
            citecolor = blue,
            anchorcolor = blue]{hyperref}
\usepackage[capitalise]{cleveref}
\usepackage[table]{xcolor}
\usepackage{makecell}

\newcommand{\approptoinn}[2]{\mathrel{\vcenter{
  \offinterlineskip\halign{\hfil$##$\cr
    #1\propto\cr\noalign{\kern2pt}#1\sim\cr\noalign{\kern-2pt}}}}}

\usepackage{float}
\usepackage{flushend}
\usepackage{comment}
\definecolor{earthgreen}{RGB}{230,255,230}
\definecolor{marsred}{RGB}{255,230,230}
\definecolor{moongrey}{RGB}{240,240,240}
\definecolor{superblue}{RGB}{210,240,255}

\title{Toward Lunar Legged Robots: Field Deployment Lessons at LUNA}
\author{Adrian Fuhrer, Joseph Church, Oliver Fischer, William Talbot, Nicolas Faesch, Yannic Hofmann, \\ Hendrik Kolvenbach, Yusuke Tanaka, and Marco Hutter% <-this % stops a space
}

\begin{document}

% This allows you to play around with how the references show up in the paper -- can be used to make space :))
\bstctlcite{IEEEexample:BSTcontrol}

\twocolumn[{%
\renewcommand\twocolumn[1][]{#1}%
    \maketitle
    \begin{center}
        \centering
        \includegraphics[width=0.99\textwidth,trim={0cm 0cm 0cm 0cm},clip]{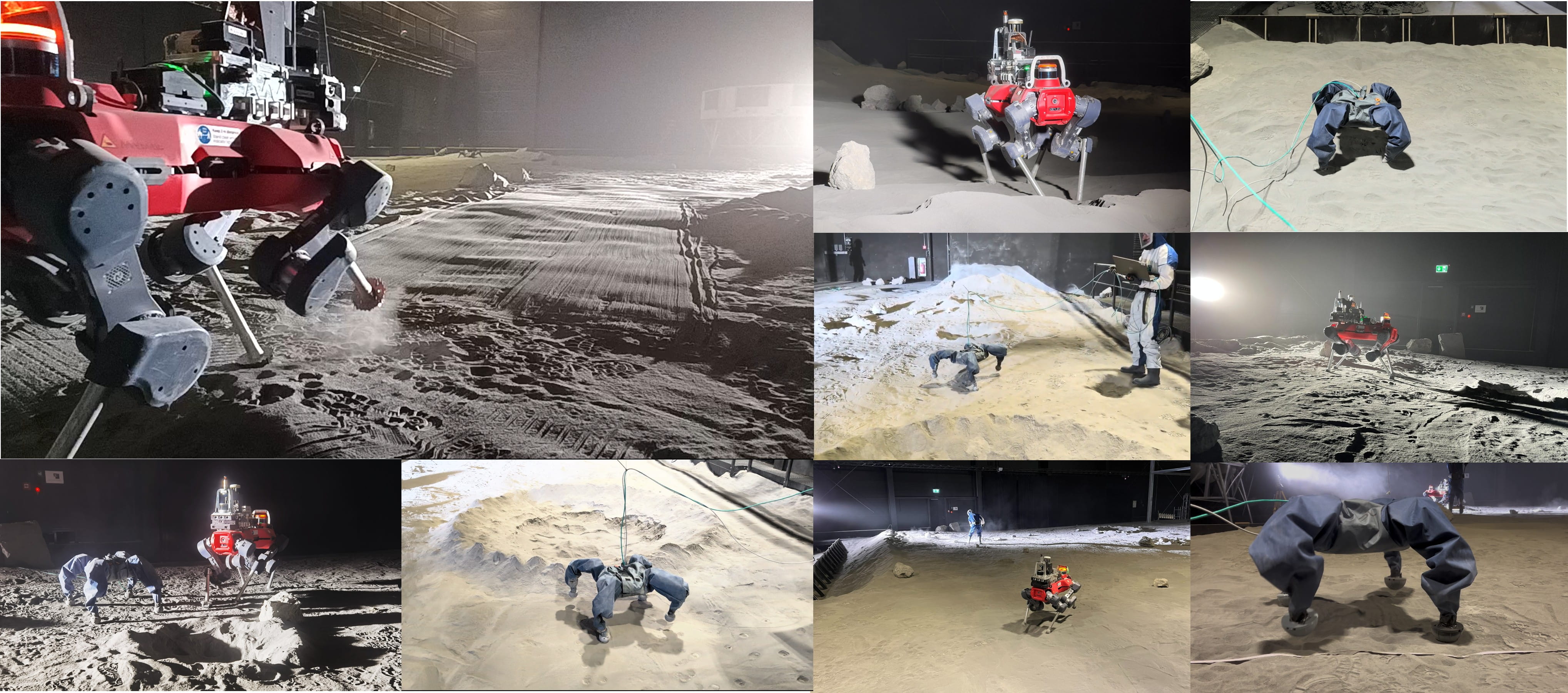} % 
    \captionof{figure}{LUNA testing campaign in 2025.}
    \label{fig:fig1}
    \end{center}%
    }]
    \footnotetext[1]{All authors are with ETH Zurich, Robotics Systems Lab; Leonhardstrasse 21, 8092 Zurich, Switzerland.
        Contact: {\tt\small yutanaka@ethz.ch}}%
\thispagestyle{empty}
\pagestyle{empty}

%%%%%%%%%%%%%%%%%%%%%%%%%%%%%%%%%%%%%%%%%%%%%%%%%%%%%%%%%%%%%%%%%%%%%%%%%%%%%%%%
\begin{abstract}
Legged robots are promising candidates for future lunar surface missions because they can traverse steep, loose, and obstacle-rich terrain that challenges conventional wheeled rovers. However, readiness for lunar deployment is limited by uncertainties in foot-regolith interaction, dust generation, illumination-driven perception degradation, and operational constraints. This paper reports lessons from the 2025 LUNA analogue campaign, where ANYmal-D and Magnecko traversed loose regolith simulant and crater-like terrain and collected long-horizon navigation and visual-inertial data under challenging lighting. We show that quadrupedal robots can traverse regolith simulant, but performance is affected by sinkage and slip, dust-generating contacts, and perception failures caused by overexposure, shadows, and low-texture regions. These results motivate tighter integration of regolith-aware locomotion policies, illumination-robust perception, repeatable analogue testing, and mission-level operational validation for future lunar legged robots.
\end{abstract}

%%%%%%%%%%%%%%%%%%%%%%%%%%%%%%%%%%%%%%%%%%%%%%%%%%%%%%%%%%%%%%%%%%%%%%%%%%%%%%%%

\section{INTRODUCTION}

Planetary exploration is entering a renewed phase of activity~\cite{nasa2025artemisIII}, with increasing interest in sustained surface operations~\cite{krawciw2026lunar}, infrastructure construction~\cite{uno2025moonbot, brinkmann2020enhancement}, and scientific access to challenging terrain~\cite{kolvenbach2025lunarleaper}. While wheeled rovers remain central to planetary mobility, their traversability is limited in steep, loose, or obstacle-rich terrain \cite{kolvenbach2022traversing}. This limitation is particularly relevant on the Moon, where loose regolith and difficult-to-access regions motivate alternative mobility systems~\cite{kolvenbach2022traversing, kolvenbach2025lunarleaper}.

Legged robots are increasingly being considered for future lunar surface operations \cite{kolvenbach2021quadrupedal} because they can actively adapt contacts, step over obstacles, and traverse terrain that may be inaccessible to wheeled platforms~\cite{bartsch2012spaceclimber, kolvenbach2025lunarleaper,spacebok}. However, despite recent progress in agile locomotion, perception, and navigation for terrestrial quadrupeds, a gap remains in understanding which improvements are critical for deployment on the Moon. In particular, open questions remain in how learning-based locomotion policies should account for regolith interaction, how foot designs could affect traction, sinkage, and dust generation, and how perception and navigation stacks should handle low-texture terrain, strong shadows, and changing illumination.

Analogue testing provides an important intermediate step toward addressing these questions. Lunar analogue testing campaigns allow robotic systems to be evaluated under more realistic terrain, operational, and integration constraints than typical laboratory settings \cite{uno2025moonbot, schuster2020arches}. Such campaigns are especially valuable for identifying not only locomotion limitations, but also system-level bottlenecks involving terrain interaction, perception, navigation, and deployment logistics.

% This paper presents lessons learned from the LUNA testing campaign, in which the quadrupedal robots, ANYmal-D and Magnecko~\cite{leuthard2024magnecko}, were deployed in the ESA/DLR LUNA lunar analogue facility~\cite{casini2020lunar}. This paper presents the tested scenarios, observed robot behaviors, lessons learned, and design implications for future lunar legged robots. Through these observations, we identify remaining key technology gaps toward higher-readiness legged robotic mobility on the Moon.

This paper reports system-level deployment lessons from operating two quadrupedal robots, ANYmal-D~\footnote{https://www.anybotics.com/robotics/anymal/} and Magnecko~\cite{leuthard2024magnecko}, in the ESA/DLR LUNA lunar analogue facility~\cite{casini2020lunar}. The testing campaign was conducted in 2025 and identifies practical gaps that emerge when legged robotic systems are operated on lunar-regolith simulant under facility-scale terrain, lighting, sensing, and operational constraints. These observations provide insight into how regolith interaction, dust generation, illumination-dependent perception, and field logistics affect the readiness of legged robots for future lunar surface missions.

The main contributions of this paper are:
\begin{itemize}
\item field deployment observations from two quadrupedal robots operating in the LUNA lunar analogue facility.
\item analysis of key system-level challenges in regolith locomotion, perception, dust exposure, and field operations.
\item lessons and design implications for future lunar legged robots and repeatable analogue testing.
\end{itemize}

\section{Background and Campaign Context}
Legged robots have been investigated for space and planetary exploration because they have the potential to traverse steep \cite{bartsch2012spaceclimber, spacebok}, irregular, and discontinuous terrain \cite{tanaka2025scaler} that is difficult for conventional wheeled rovers. Legged and limbed systems have demonstrated walking, hopping \cite{spiridonov2024spacehopper, tanaka2025tethered}, climbing \cite{tanaka2025scaler,uno2026limbero}, and contact-rich mobility \cite{shirai2022simultaneous, uno2025moonbot} for planetary scenarios.

Recent progress in agile and learning-based legged locomotion has further improved the robustness of quadrupedal robots in complex terrestrial environments \cite{miki2022elevation}. However, for lunar applications, these controllers must also address reduced gravity~\cite{arm2025efficient}, limited power and thermal budgets, and interaction with loose regolith. Learning-based locomotion policies are still trained with simplified contact models \cite{choi2023learning}, leaving gaps in understanding their behavior on granular terrain and dust generation.

Lunar analogue campaigns bridge simulation, lab testing, and mission deployment \cite{schuster2020arches}, revealing system-level issues in locomotion, perception, navigation, communication, and operations \cite{arm2023scientific, ligeza2026semi}. Key questions remain on regolith interaction for RL policies, foot design trade-offs, and perception in low-texture, high-contrast, changing illumination. The LUNA testing campaign addressed these gaps using ANYmal-D and Magnecko in a controlled lunar analogue facility.

\section{Campaign Overview and Experimental Setup}
\subsection{LUNA Analogue facility and site setup}
\begin{figure}
    \centering
    \begin{subfigure}[t]{0.49\linewidth}
        \centering
        \includegraphics[width=\linewidth]{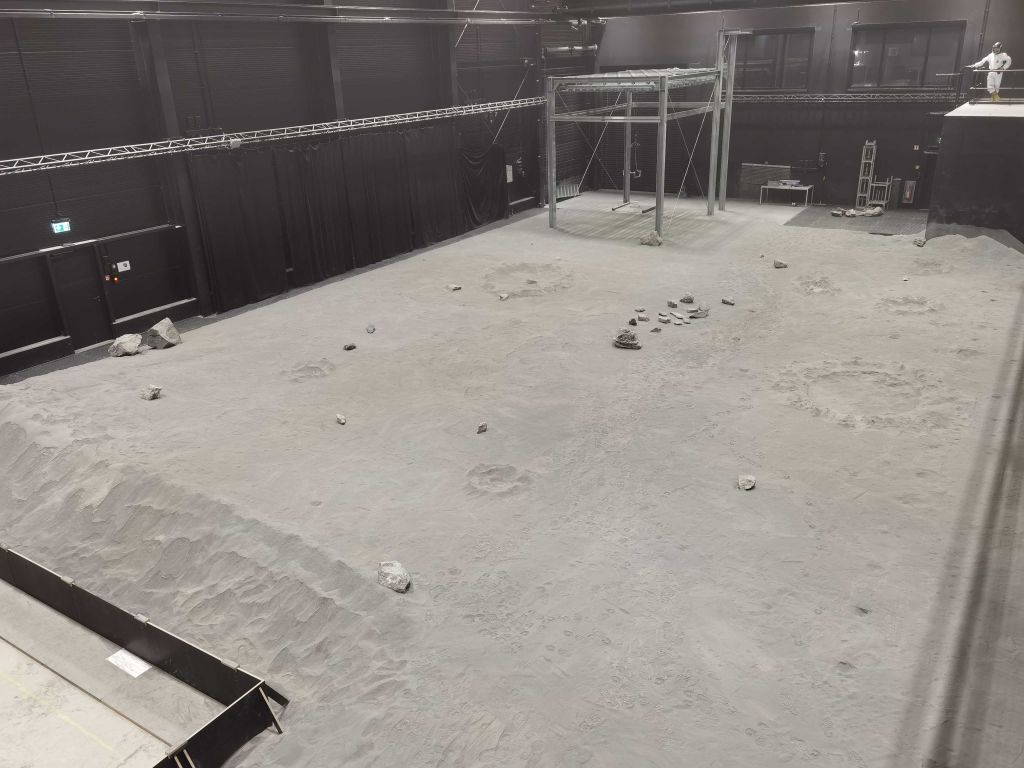}
        \caption{High-angle illumination (only one row on during testing.}
        \label{fig:luna-setup-high}
    \end{subfigure}\hfill
        \begin{subfigure}[t]{0.49\linewidth}
        \centering
        \includegraphics[width=\linewidth]{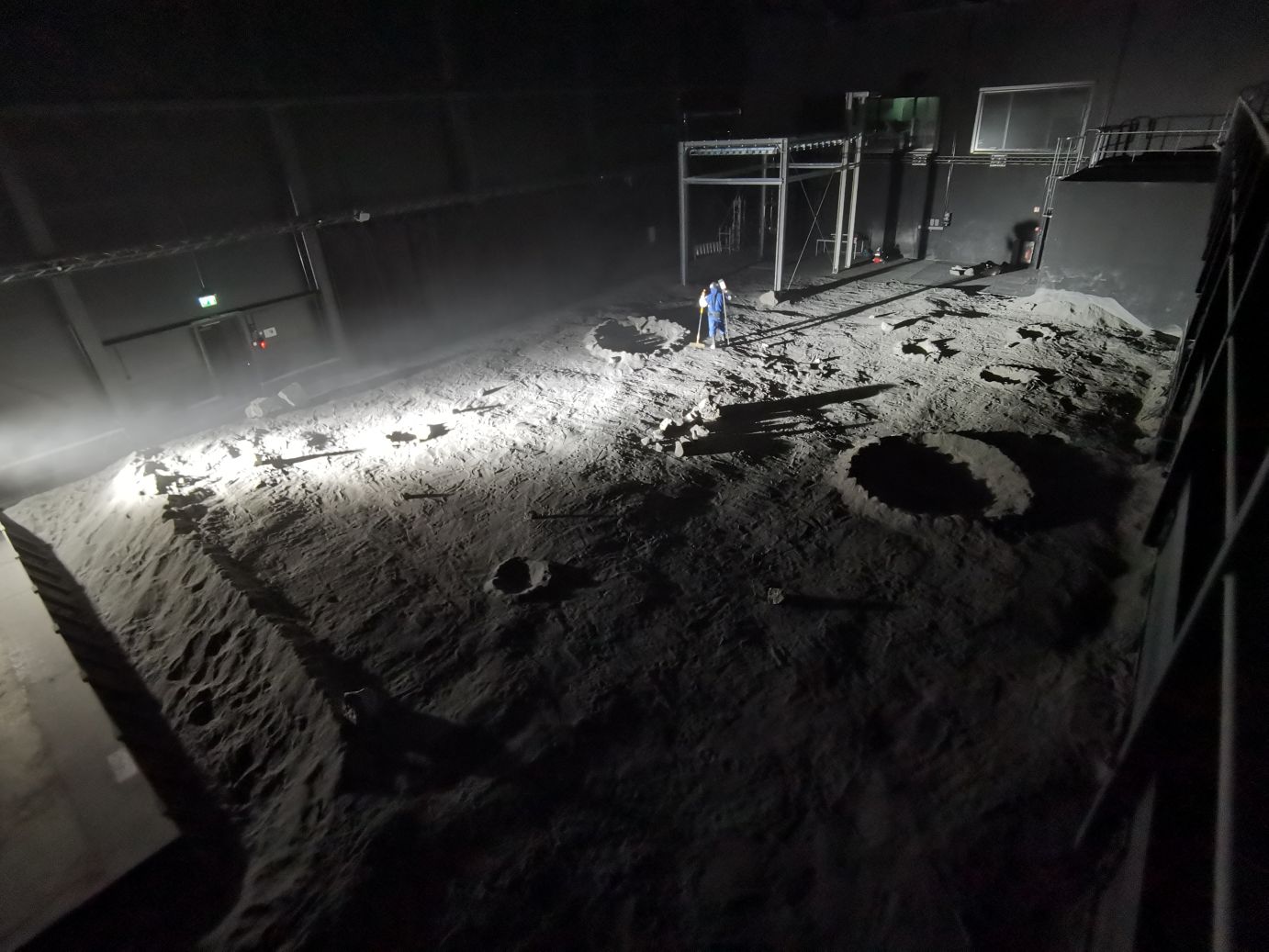}
        \caption{Low-angle illumination}
        \label{fig:luna-setup-low}
    \end{subfigure}
    \caption{LUNA regolith field configuration and lighting conditions used during the campaign. The setup includes flat, loose regolith, craters, and obstacles.}
    \label{fig:luna-setup}
\end{figure}

The LUNA lunar analogue facility (ESA/DLR) at the DLR site in Cologne, Germany, provides a controlled indoor environment for testing lunar surface mobility and operations~\cite{casini2020lunar}. The main test area is an approximately \SI{700}{\square\meter} regolith field filled with EAC-1A simulant, with controllable lighting (blackout curtains and adjustable illumination) to emulate different lunar illumination conditions.
%Compared with standard laboratory floors or small-scale terrain boxes, the LUNA facility allows full robotic systems to be tested over larger distances and in more realistic terrain configurations. 

The regolith bed was configured to include flat regions, craters, and obstacle areas as shown in \cref{fig:luna-setup}. The navigation stack was evaluated under both low- (\cref{fig:luna-setup-low}) and high-angle (\cref{fig:luna-setup-high}) illumination conditions. Although the regolith simulant EAC-1A is not fully representative of lunar dust behavior, its very low water content and dry nature provide a unique opportunity to gain insight.

\subsection{Hardware setup}
\subsubsection{Magnecko}
\begin{figure*}[t]
    \centering
    \begin{subfigure}[t]{0.32\textwidth}
        \centering
        \includegraphics[width=\linewidth]{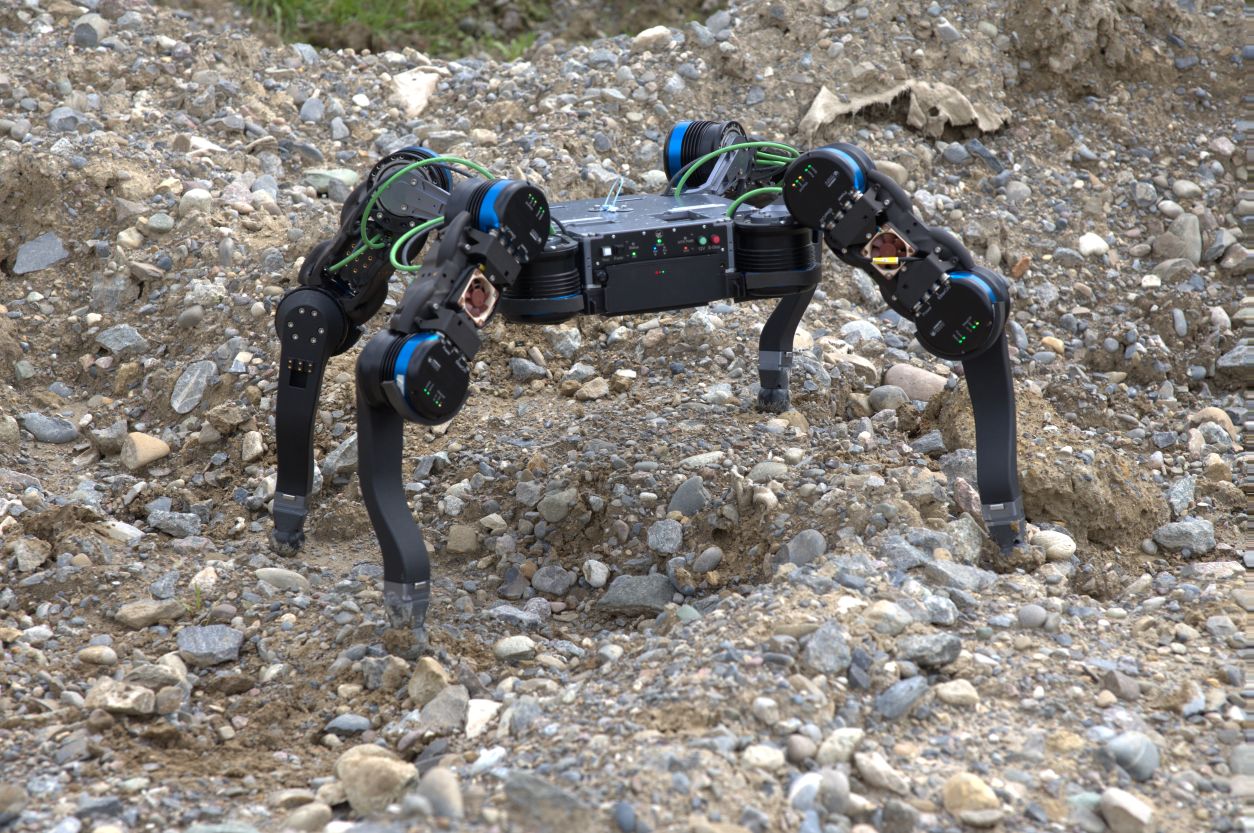}
        \caption{Magnecko hardware.}
        \label{fig:magnecko-render}
    \end{subfigure}\hfill
    \begin{subfigure}[t]{0.32\textwidth}
        \centering
        \includegraphics[width=\linewidth]{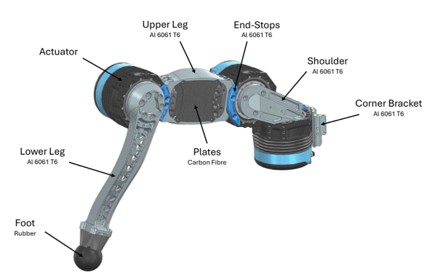}
        \caption{Magnecko leg configuration.}
        \label{fig:magnecko-leg}
    \end{subfigure}\hfill
    \begin{subfigure}[t]{0.32\textwidth}
        \centering
        \includegraphics[width=\linewidth]{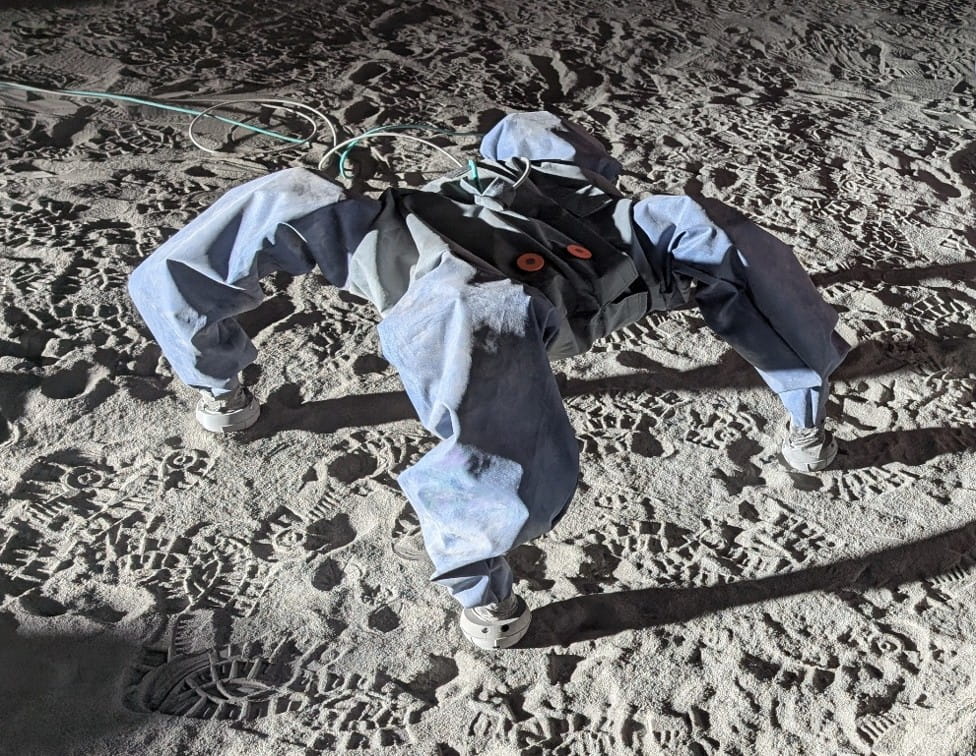}
        \caption{Magnecko with the dust suit.}
        \label{fig:magnecko-suit}
    \end{subfigure}

    \caption{Magnecko, a quadruped robot used for the LUNA testing campaign.}
    \label{fig:magnecko}
\end{figure*}

Magnecko~\cite{leuthard2024magnecko}, shown in \cref{fig:magnecko-render}, is a quadrupedal robot developed in-house for climbing and rough-surface locomotion. The platform has a mass of \SI{16}{\kilogram} and is equipped with onboard computation and battery power for untethered operation. Its modular architecture allows rapid modification of foot modules, sensor suite, payload mounts, and test instrumentation.

The robot has four legs with three degrees of freedom in an insect-like configuration (\cref{fig:magnecko-leg}). Each joint uses a custom actuator with a compound planetary gearbox, providing \SI{45}{\newton\meter} peak torque and \SI{29}{\radian\per\second} no-load speed. The morphology is omnidirectional and enables torso-height adjustment useful on uneven or sloped terrain.

% Since Magnecko was not designed for lunar or dusty environments, the robot's joints and torso are not fully protected against regolith simulant ingress. For this LUNA campaign, the robot was therefore wrapped in a custom dust-protection suit, as shown in Fig.~\ref{fig:magnecko-suit}. This suit was intended as a temporary campaign-level protection measure rather than a permanent design solution. In particular, such coverings may obstruct heat dissipation, limit the openings for sensors and mechanisms, and raise durability concerns for longer-duration operation. Nevertheless, the suit provided sufficient protection in this campaign.

Magnecko's joints and torso are not sufficiently sealed against regolith simulant ingress. The robot was therefore wrapped in a temporary dust-protection suit, as shown in \cref{fig:magnecko-suit}. While sufficient for short-duration testing, such coverings can limit heat dissipation and long-term durability, as we discovered for the joints on ANYmal-D.

% \subsubsection{Gravity Offloading Mechanisms}
% To investigate the effect of reduced gravity on legged locomotion over loose regolith simulant, Magnecko was also tested with a custom gravity-offloading mechanism \cite{arm2025efficient}. The mechanism applies an approximately vertical force to the robot via a tether pulled by a constant-force spring reel. 
% %. This force was generated using a constant-force spring reel and was set to compensate approximately (5/6) of the robot weight, corresponding to the difference between terrestrial and lunar gravity. 
% As a result, the robot experienced an effective normal load closer to that expected on the Moon.
% This approach does not fully reproduce lunar gravity because it only reduces the robot's apparent weight via an external vertical force. The tether can also introduce parasitic constraints and disturbances during motion. However, the setup provides a practical first-order method for studying how reduced normal loading influences legged locomotion.

\subsubsection{Gravity Offloading}

%Magnecko was tested with a custom gravity-offloading mechanism to approximate reduced normal loading during locomotion over loose regolith simulant~\cite{arm2025efficient}. The mechanism applied an approximately vertical upward force through a tether and a constant-force spring reel. Although this does not fully reproduce lunar gravity and can introduce tether-related constraints, it provides a practical first-order method for evaluating how reduced normal load affects foot sinkage, traction, and locomotion behavior.

Magnecko was tested with a custom gravity-offloading mechanism to approximate reduced normal loading during locomotion over loose regolith simulant~\cite{arm2025efficient}. The reduced-gravity approach combined an approximately vertical upward force applied through a tether and constant-force spring reel with gravity-compensated feedforward torques in the locomotion controller. Although this approach does not fully reproduce lunar gravity and can introduce tether-related constraints, it provides a practical first-order method for evaluating how reduced normal load affects foot sinkage, traction, and locomotion behavior.

\subsubsection{ANYmal-D with BOXi}

\begin{figure}[t]
    \centering
    \includegraphics[width=0.8\linewidth]{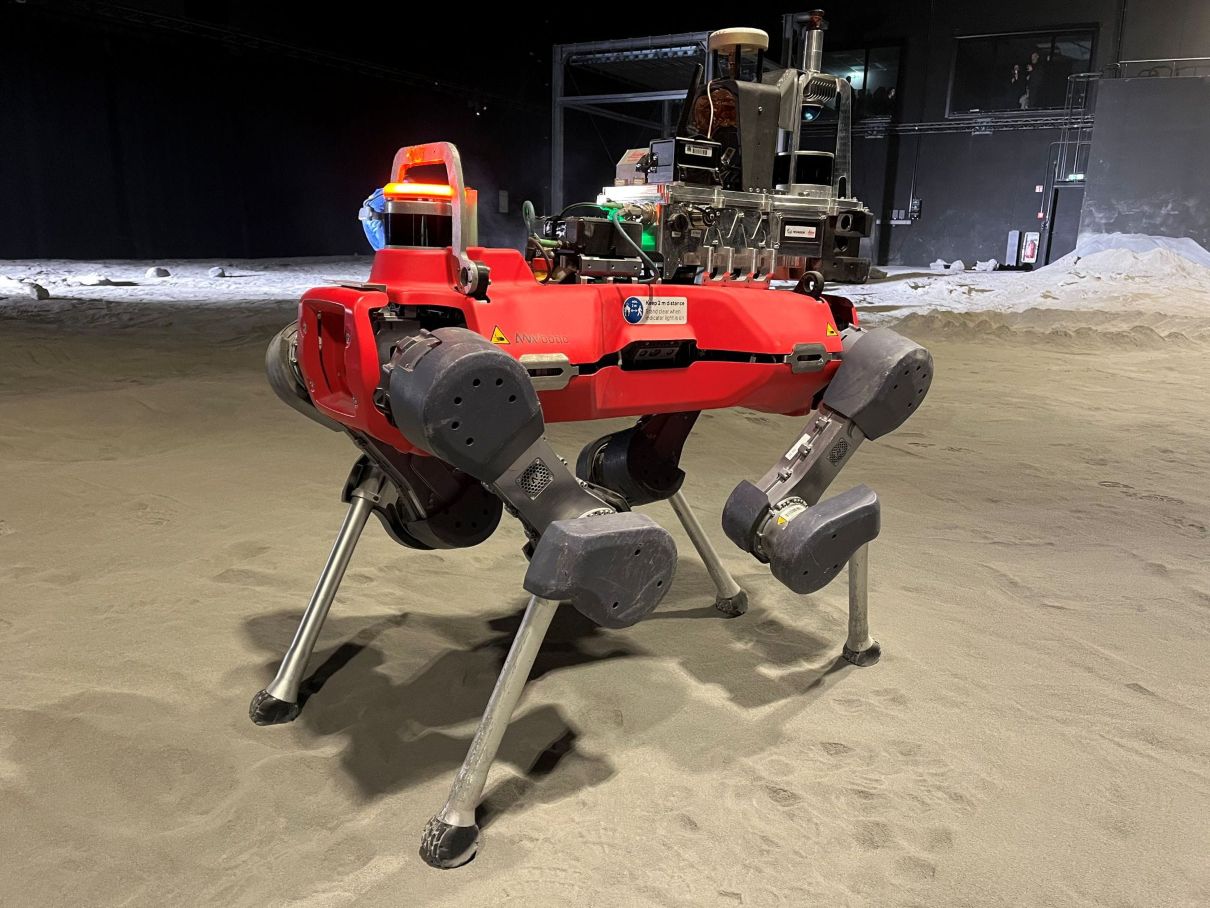}
    \caption{ANYmal-D equipped with the BOXi payload used in the LUNA testing campaign.}
    \label{fig:anymal-boxi}
\end{figure}

ANYmal-D was used for long-horizon navigation testing and equipped with the BOXi perception payload, as shown in \cref{fig:anymal-boxi}. BOXi~\cite{frey2025boxi} is a modular, multimodal sensor suite designed for mobile robot data collection and algorithm evaluation, featuring synchronized perception sensors and onboard computation for mapping, localization, and navigation experiments, and data collection during traverses in the LUNA regolith field. ANYmal-D is water- and dust-proof to IP67.

The camera was mounted at a height of \SI{70}{\centi\meter}, and the commanded walking speed was kept between \SI{0.2}{\meter\per\second} and \SI{0.6}{\meter\per\second} to capture realistic vibration from regolith interaction and viewpoint changes induced by legged locomotion.
Datasets were collected with external ground truth from a total station placed directly behind the main light source. 

%The ANYmal-D--BOXi setup provided a robust platform for collecting sensor data during traverses over the regolith simulant under different terrain and illumination conditions. The payload enabled the navigation stack to operate using onboard perception while also recording synchronized data for post-campaign analysis. In the LUNA testing campaign, this setup was used to evaluate the navigation stack's behaviour during longer traverses, terrain mapping, obstacle-rich navigation, and low- and high-angle illumination tests. 

\subsubsection{Foot end-effector attachments}
Three different foot end-effector geometries (\cref{fig:foot-end-effectors}) were used in this campaign to investigate potential benefits and disadvantages on both robots. The larger spherical foot increases the contact area (\cref{fig:foot-large}), and the high-traction foot uses flat plates with sawtooth-like teeth (\cref{fig:foot-traction}).

\subsection{Reinforcement Learning-Based Locomotion Policy}

Magnecko used an RL locomotion policy trained in simulation with an IsaacLab actor-critic pipeline using PPO \cite{IsaacLab,schulman2017proximal}. The policy tracked planar velocity commands and output joint-position targets for the low-level controllers with similar observation and reward functions as presented in \cite{arm2025efficient}, and sim-to-real used PACE for calibration \cite{bjelonic2025towards}. In simulation, contacts were modeled with rigid-body Coulomb friction, so the policy did not explicitly capture granular regolith interaction.
For ANYmal-D, the stock locomotion policy was used, as it was operated on relatively flat surfaces under Earth's gravity. 

\subsection{Regolith Simulant Surface Preparation}
The regolith simulant bed was prepared manually for the campaigns. The traversable area available for the missions was \SI{19}{\meter} by \SI{17}{\meter} with various rocks to resemble lunar-relevant terrain appearance.
%The team used this setup to evaluate several foot designs on the ANYmal platform (in part motivated by hardware constraints). 
Qualitative locomotion tests were performed on two terrain conditions representative of a lunar mission. First, the surface was loosened to a depth of $\approx\SI{200}{\milli\meter}$ to mimic a \emph{loose regolith environment}, i.e., fluffy near-surface material such as in the upper regolith layers or in an artificial crater-like feature (e.g., crater rims). Second, the surface was compacted to approximate mature soil and transitioned to rocky terrain. Throughout this paper, we refer to these conditions as \emph{loose regolith} and \emph{compact/flat regolith}, respectively.
These surface preparations are essential for evaluating dust dropout and general contact behaviour at the end-effector.

\begin{figure}
    \centering
    \includegraphics[width=0.8\linewidth]{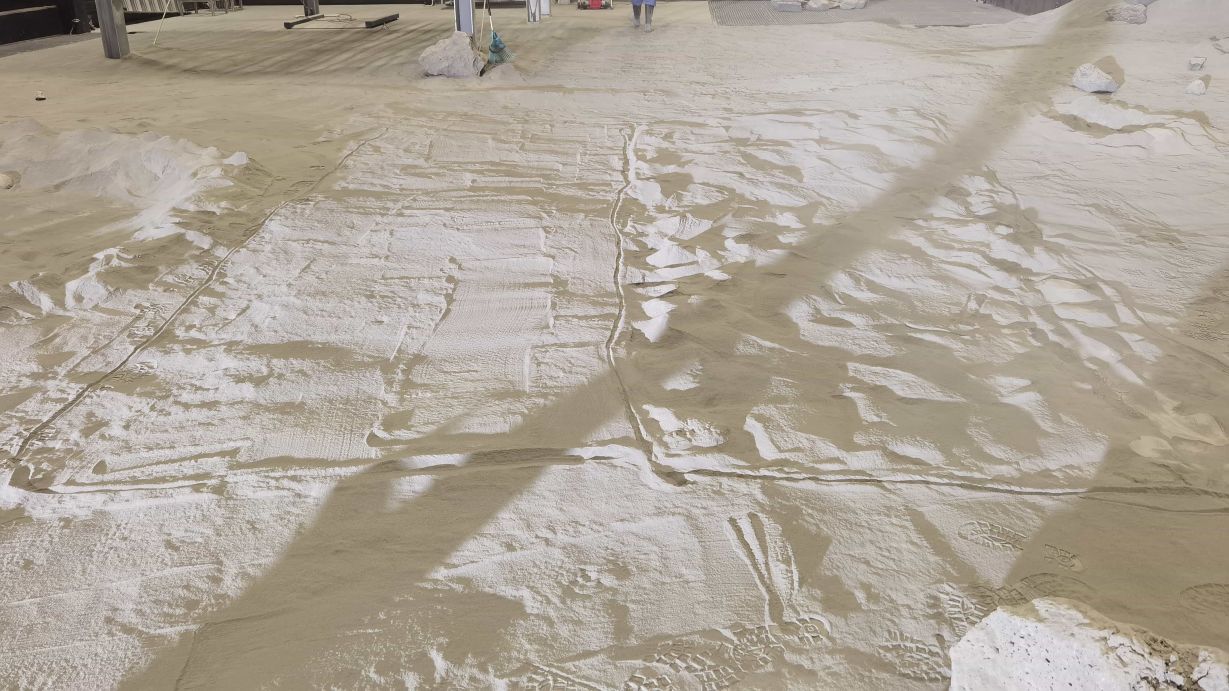}
    \caption{Preparation of the terrain to test on, loosened ground on the right, compacted soil on the left. }
    \label{fig:anymal-soil-prep}
\end{figure}

\begin{figure}[!t]
    \centering
    \begin{subfigure}[t]{0.32\columnwidth}
        \centering
        \includegraphics[width=\linewidth]{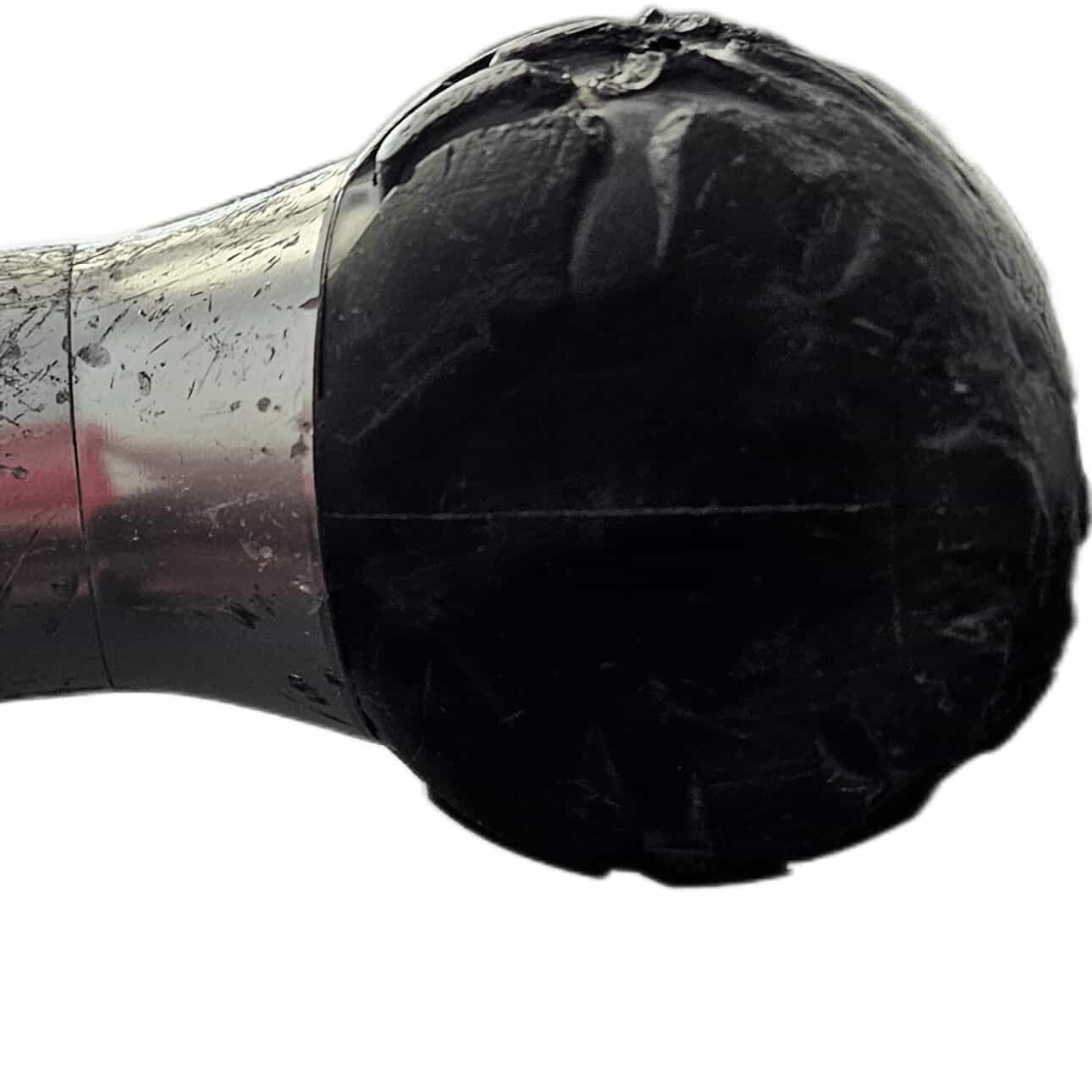}
        \caption{Stock spherical foot (\SI{65}{\milli\metre} diameter).}
        \label{fig:foot-stock}
    \end{subfigure}\hfill
    \begin{subfigure}[t]{0.32\columnwidth}
        \centering
        \includegraphics[width=\linewidth]{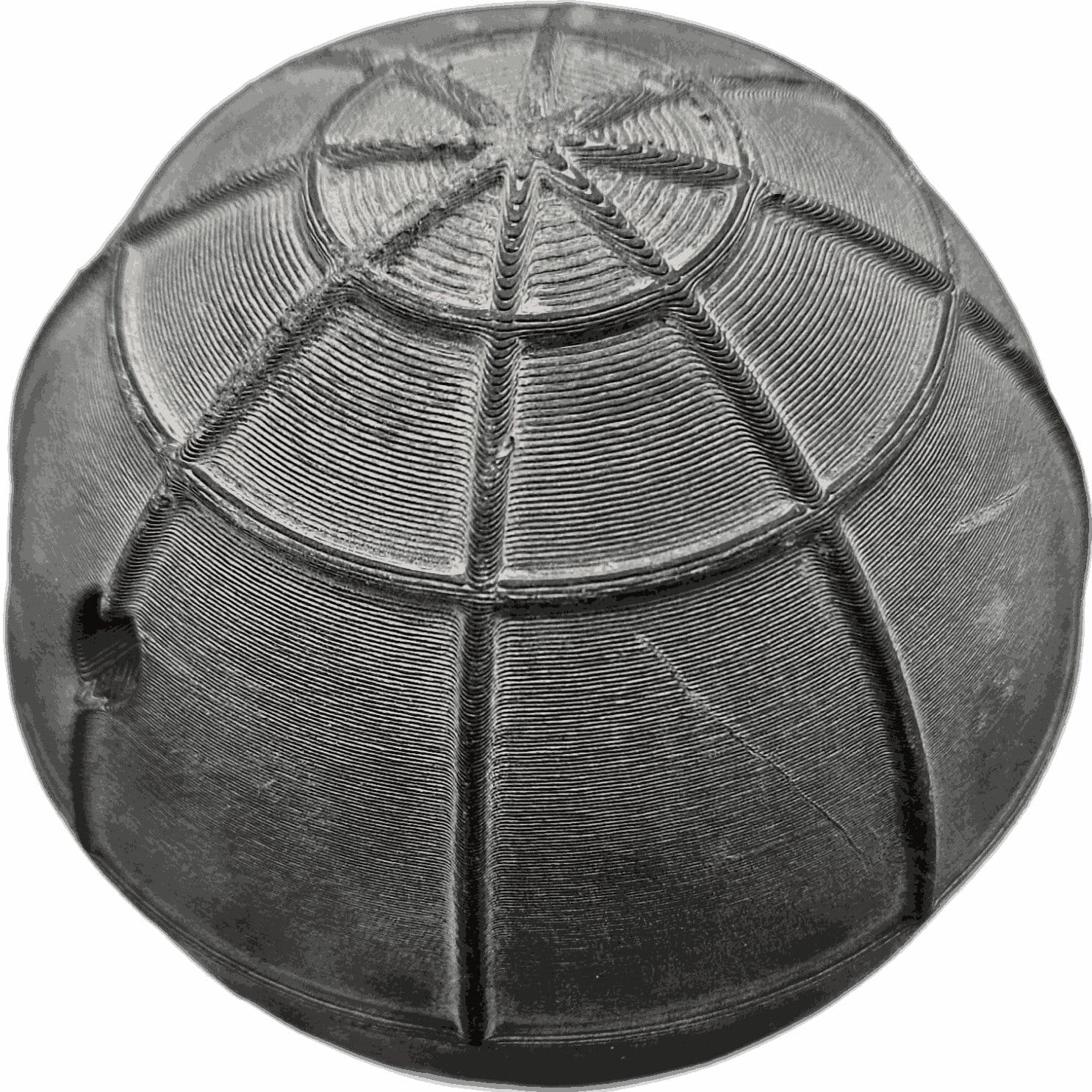}
        \caption{Large spherical foot (\SI{75}{\milli\metre} diameter).}
        \label{fig:foot-large}
    \end{subfigure}\hfill
    \begin{subfigure}[t]{0.32\columnwidth}
        \centering
        \includegraphics[width=\linewidth]{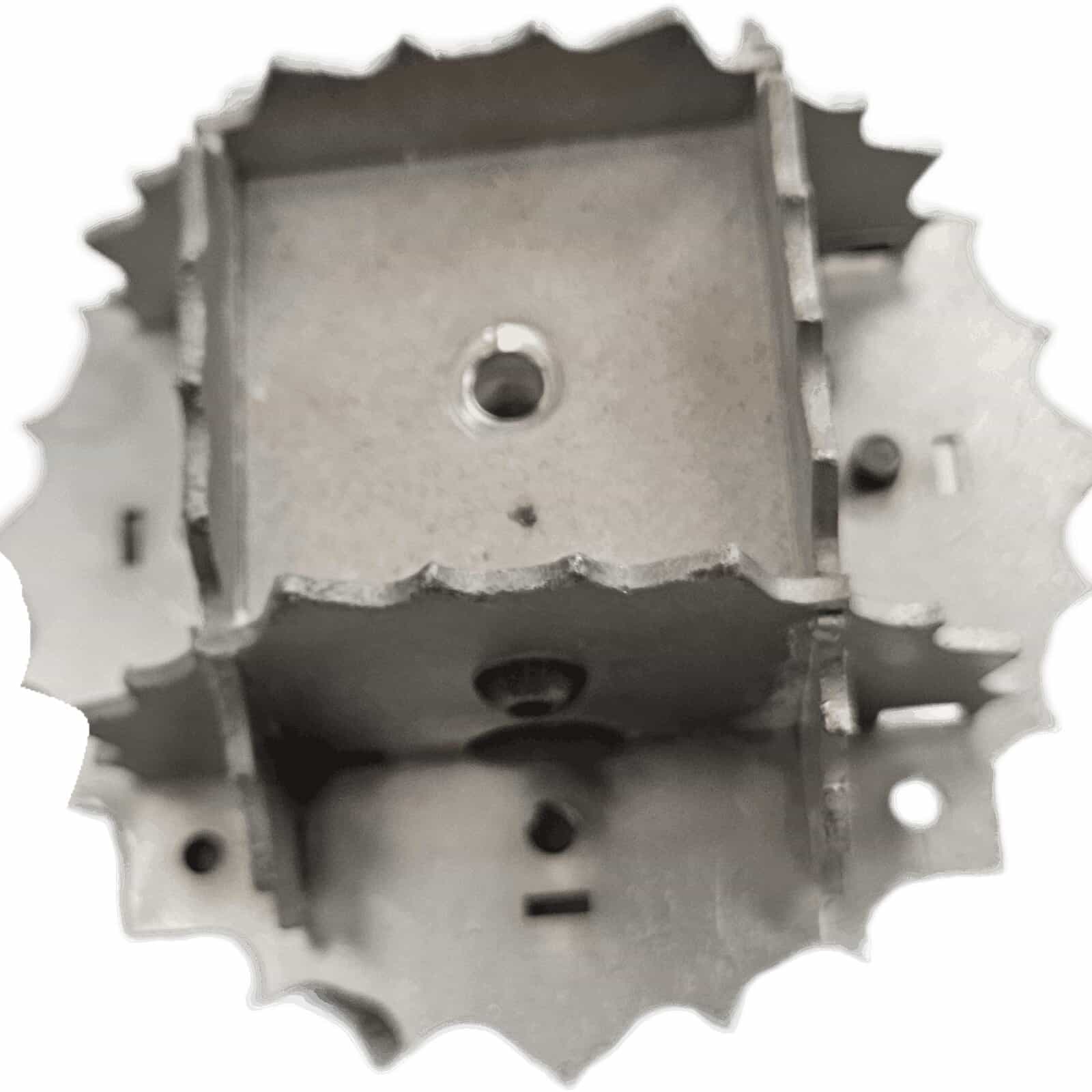}
        \caption{High-traction foot (\SI{95}{\milli\metre} diameter) with sawtooth-like edges.}
        \label{fig:foot-traction}
    \end{subfigure}
    \caption{Foot end-effector designs evaluated during the campaign.}
    \label{fig:foot-end-effectors}
\end{figure}

\begin{table*}[!t]
\centering
\footnotesize
\caption{Summary of LUNA testing campaign tests, observations, and design implications for lunar legged robots.}
\label{tab:campaign-summary}
\renewcommand{\arraystretch}{1.15}
\begin{tabular}{p{0.14\textwidth} p{0.08\textwidth} p{0.16\textwidth} p{0.25\textwidth} p{0.25\textwidth}}
\toprule
\textbf{Test theme} & \textbf{Platform} & \textbf{Scenario} & \textbf{Key observation} & \textbf{Design implication} \\
\midrule
Loose regolith locomotion\newline(\cref{sec:loose-regolith}) & Magnecko & Loose crater-like terrain with gravity offloading & The robot traversed crater-like terrain, but foot sinkage, slip, and reduced traction required command adaptation. & Locomotion policies should include regolith sinkage, slip, and terrain deformation effects. \\
\midrule
Foot end-effector comparison\newline(\cref{sec:foot-endeffector}) & Magnecko, ANYmal-D & Stock, enlarged, and high-traction foot geometries on prepared regolith & Terrain preparation and local packing had stronger effects than foot geometry in the qualitative trials. & Foot designs require more repeatable regolith-bed testing before drawing quantitative conclusions. \\
\midrule
Long-horizon navigation\newline(\cref{sec:long-horizon-nav}) & ANYmal-D & 11 missions over a \SI{19}{\meter} by \SI{17}{\meter} traversable area & A long-horizon, multimodal navigation dataset capturing realistic regolith-induced vibration, viewpoint, and perception degradation. & The key outcome is a dataset that can be reused as a benchmark for future lunar navigation stacks under realistic locomotion and lighting. \\
%Illumination-dependent perception\newline(\cref{sec:long-horizon-nav}) & ANYmal-D & Raking spotlight-like illumination and partial ceiling lighting & Overexposure, shadows, low texture, and dust in the field of view degraded visual feature tracking. & Lunar navigation requires illumination-aware exposure control, shutter selection, and robust visual-inertial perception. \\
\midrule
Illumination-dependent perception\newline(\cref{sec:long-horizon-nav}) & ANYmal-D & Raking spotlight-like illumination and partial ceiling lighting & Overexposure, shadows, low texture, and dust in the field of view degraded visual feature tracking. & Lunar navigation requires illumination-aware exposure control, shutter selection, and visual-inertial perception. \\
\midrule
Dust and regolith disturbance\newline(\cref{sec:locomotion-dust}) & Magnecko, ANYmal-D & Foot drag, touchdown, and high-traction foot contact & Dust was generated by both tangential foot drag and vertical foot impact; high-traction feet produced larger visible dust clouds. & Dust generation should be treated as a locomotion RL policy constraint, not only an environmental protection problem. \\
\midrule
Field operations and repeatability\newline(\cref{sec:ops-lessons}) & Magnecko, ANYmal-D & Indoor analogue facility operation & Safety limits, masking needs, unreliable WiFi, and manual terrain preparation affected repeatability. & Future campaigns need standardized terrain preparation, communication infrastructure, and operational protocols. \\
\bottomrule
\end{tabular}
\end{table*}

\subsection{Illumination setup}
The lighting conditions were varied to expose the navigation stack to challenging visual conditions expected in lunar-like environments. Most missions used a raking spotlight-like illumination configuration. For selected missions, an additional row of ceiling lights with unknown spectral properties was enabled. The controllable CCT lighting was operated at 90-100\% intensity, with color temperatures between \SI{5600}{\kelvin} and \SI{5900}{\kelvin} and a green-magenta offset between $-0.25$ and $0$. 
These settings produced strong shadows, low-texture regions, and high-contrast image areas that affected feature tracking and visual localization.

%\subsection{\red{Autonomy level and operator involvement.}}

\section{Campaign Observations and Lessons Learned}
\cref{tab:campaign-summary} summarizes the tests and key observations.

\subsection{Locomotion over Loose Regolith Simulant Crater\label{sec:loose-regolith}}

Locomotion over loose regolith crater is challenging and hazardous for rovers, since once all the wheels start slipping, there may be no way to escape \cite{ishigami2009slope}. 
Magnecko was able to walk over relatively large and small craters in the field, as shown in \cref{fig:magnecko-crater,fig:luna-mini-crater}. For the larger crater, both front legs in \cref{fig:magnecko-crater} at $t={13,15}$ were submerged to the edge of the crater, while the back two legs were slipping. Nonetheless, Magnecko reached the rim of the crater.
For the smaller crater (approximately the robot's size), only one leg was inside the crater at a time. Magnecko was able to traverse the mini crater, but the leg inside struggled to gain traction due to the large height difference across the surface and reduced gravity. This behavior may also be an artifact of the gravity-offloading mechanism used, which constrains tipping because the tether is not necessarily well aligned with the gravitational force.
For both cases, the robot was teleoperated, and the commands had to be adjusted at $t=13$~s in \cref{fig:magnecko-crater} to wiggle out of the loose regolith. Because the policy was not trained with sinkage in simulation, it struggled on the loose regolith slope.
% Incorporating these interaction effects into the simulation could improve behavior in future work.

\subsection{Foot End-Effector Designs and On-Regolith Simulant Locomotion\label{sec:foot-endeffector}}
\begin{figure*}[!ht]
        \centering
        \includegraphics[width=\linewidth]{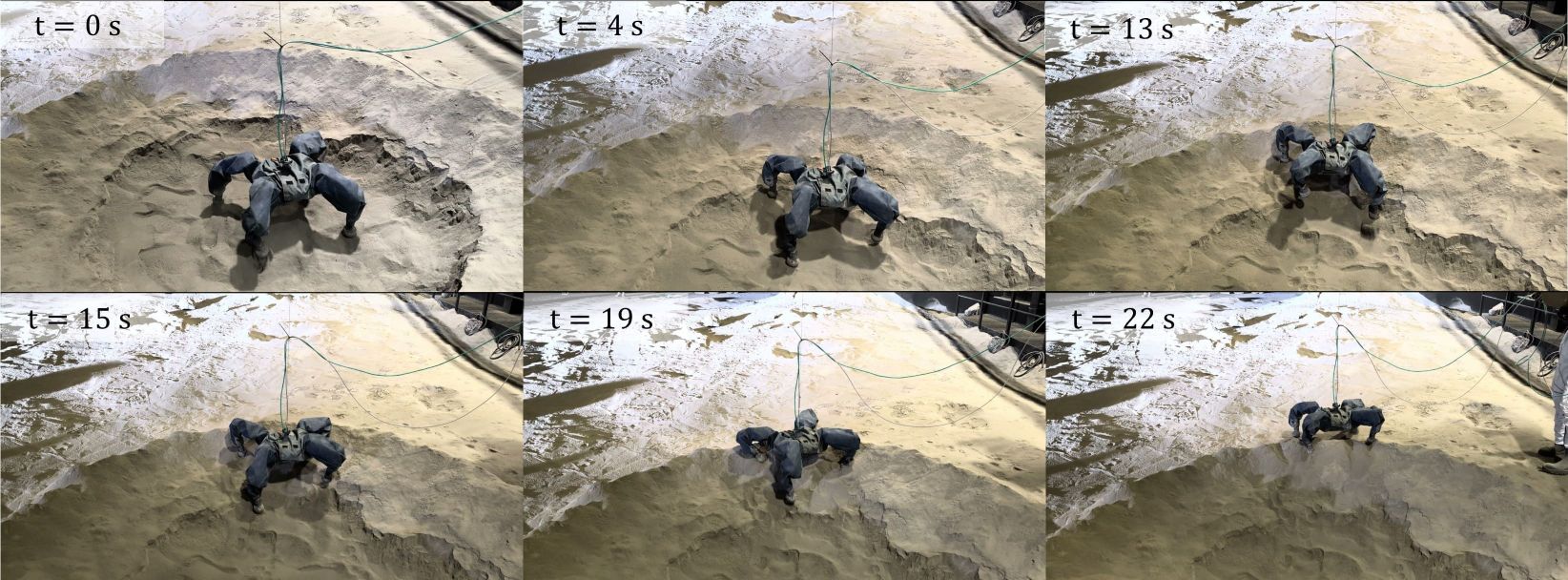}
    \caption{Magnecko walking up the crater slope from the inside.}
    \label{fig:magnecko-crater}
\end{figure*}

\begin{figure*}[t]
    \centering

    \begin{subfigure}[t]{0.32\textwidth}
        \centering
        \includegraphics[width=\linewidth]{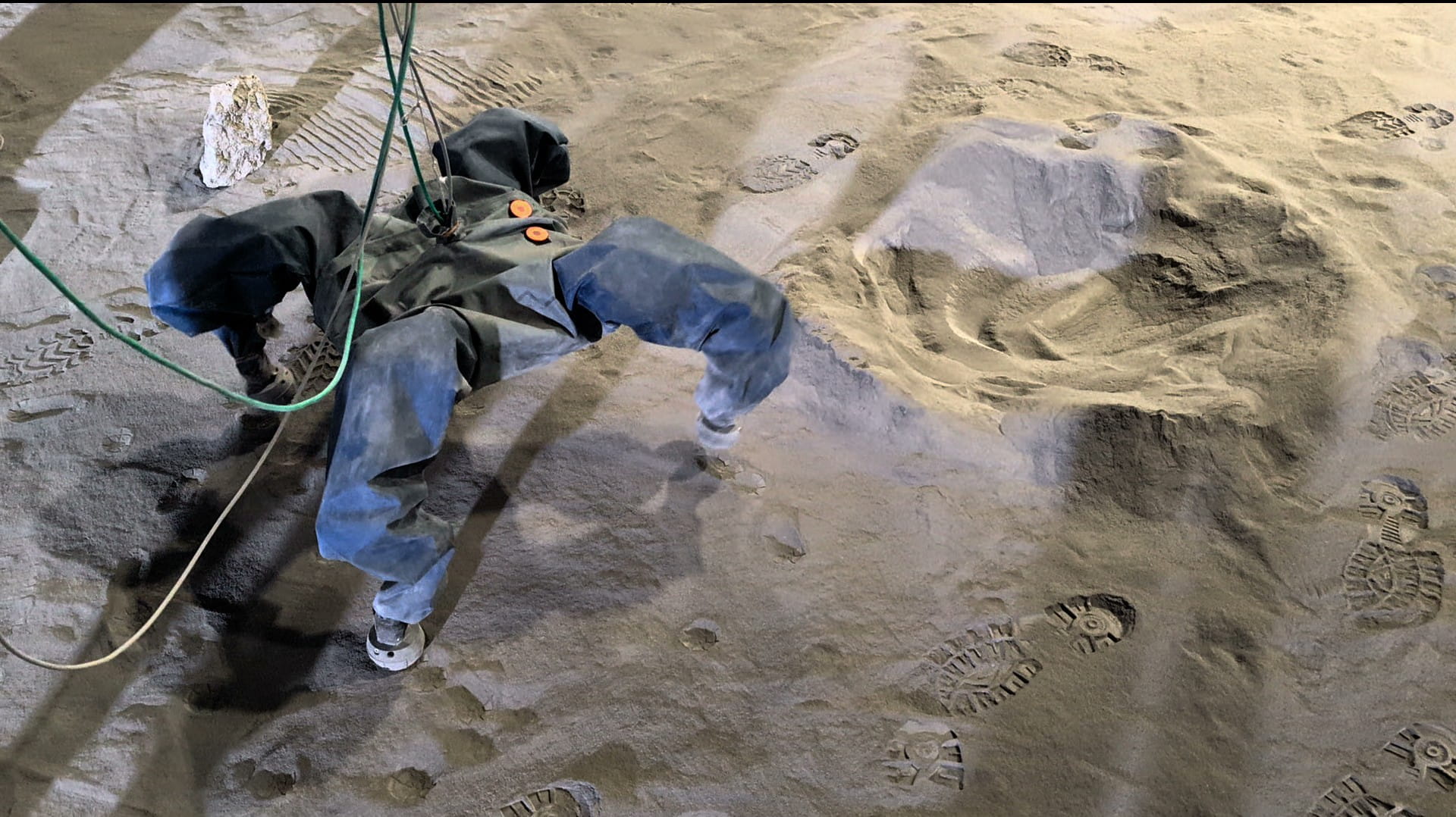}
        \caption{Magnecko walking toward the mini crater.}
        \label{fig:luna-mini-crater-approach}
    \end{subfigure}\hfill
    \begin{subfigure}[t]{0.32\textwidth}
        \centering
        \includegraphics[width=\linewidth,trim=0 5mm 0 0,clip]{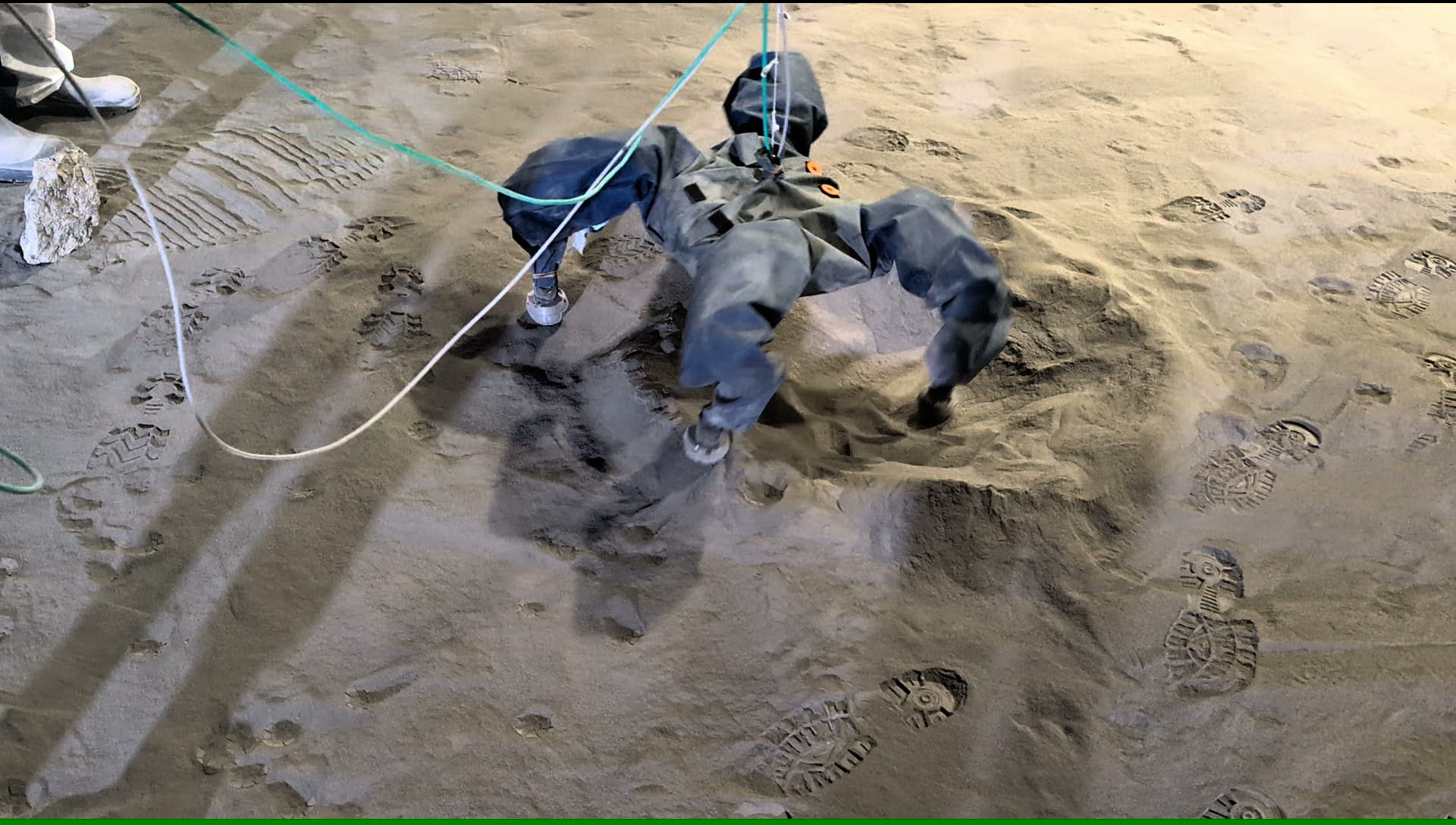}
        \caption{Magnecko walking in the mini crater.}
        \label{fig:luna-mini-crater-inside}
    \end{subfigure}\hfill
    \begin{subfigure}[t]{0.32\textwidth}
        \centering
        \includegraphics[width=\linewidth]{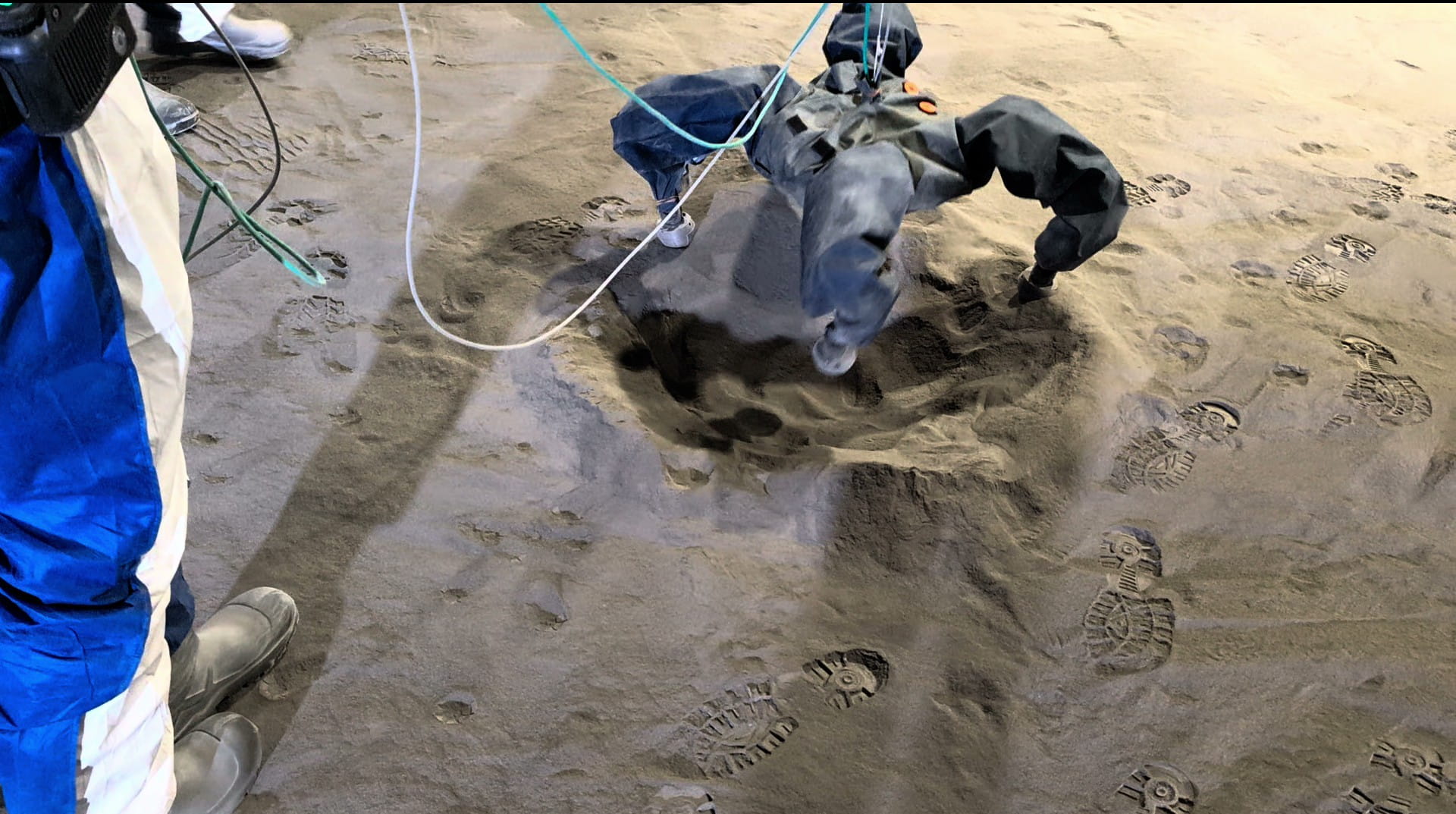}
        \caption{Magnecko walking over the mini crater.}
        \label{fig:luna-mini-crater-air}
    \end{subfigure}

    \caption{Magnecko locomotion over the mini crater with gravity offloading.}
    \label{fig:luna-mini-crater}
\end{figure*}

Across different foot geometries, we observed limited qualitative or quantitative performance differences among the three in this campaign. A larger foot can still be submerged in deep, loose regolith as shown in \cref{fig:magnecko-crater}, and in our testing, variations in the regolith simulant conditions, such as packing, had larger effects than the foot end-effector itself. Hence, foot designs need testing in a more controlled and repeatable regolith simulant bed. We did observe qualitative differences in dust generation, which is discussed in Section~\ref{sec:locomotion-dust}.

\subsection{Long-Horizon Locomotion and Navigation\label{sec:long-horizon-nav}}
% \begin{figure*}
%     \centering
%     \includegraphics[width=0.9\linewidth]{luna/mission.jpg}
%     \caption{Navigation and data collection mission trajectory overviews.}
%     \label{fig:anymal-navigation}
% \end{figure*}

\begin{figure*}[!ht]
    \centering

    % Row 1: 4 images
    \begin{subfigure}[t]{0.24\textwidth}
        \includegraphics[width=\linewidth]{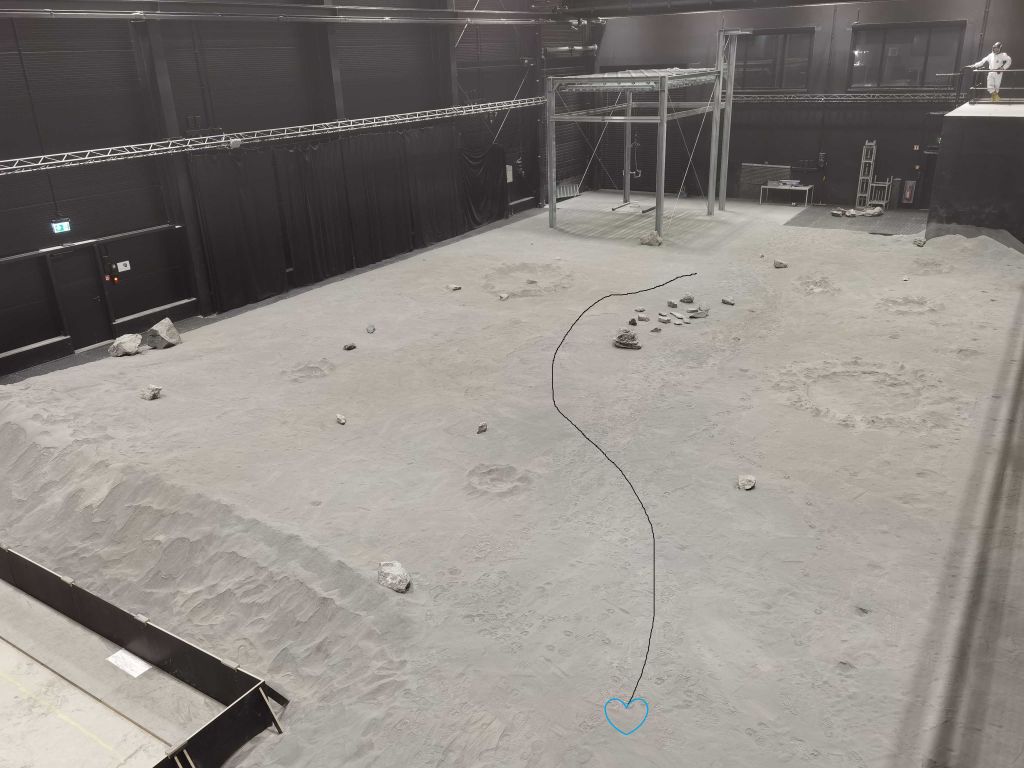}
        \caption{}
    \end{subfigure}\hfill
    \begin{subfigure}[t]{0.24\textwidth}
        \includegraphics[width=\linewidth]{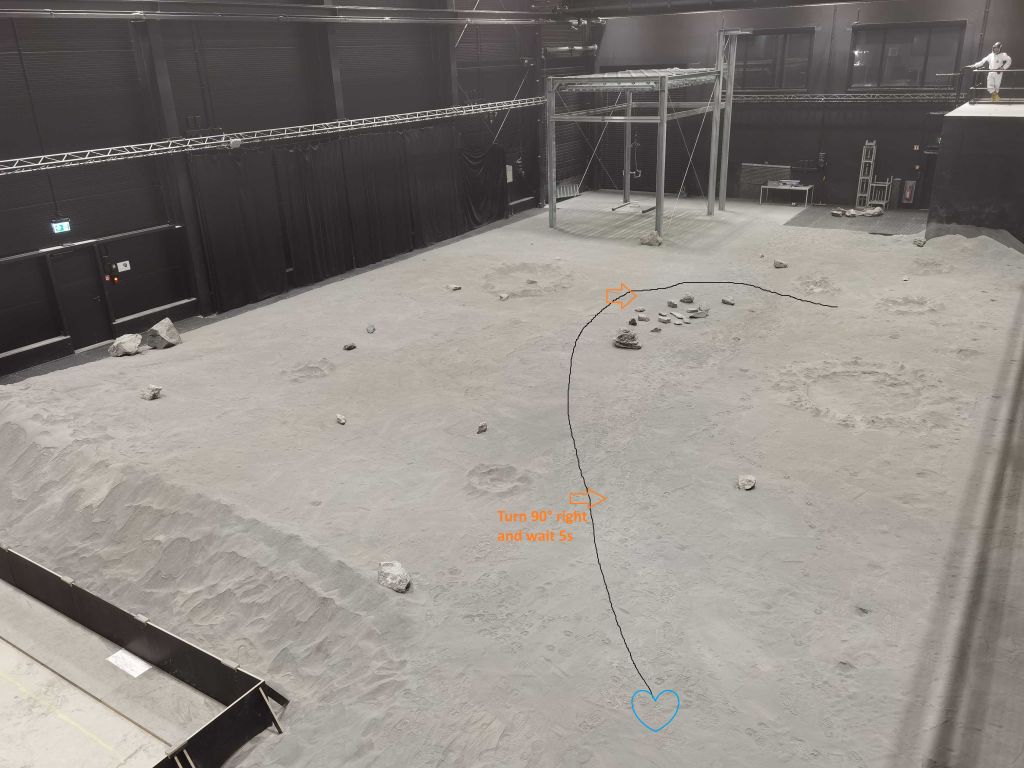}
        \caption{}
    \end{subfigure}\hfill
    \begin{subfigure}[t]{0.24\textwidth}
        \includegraphics[width=\linewidth]{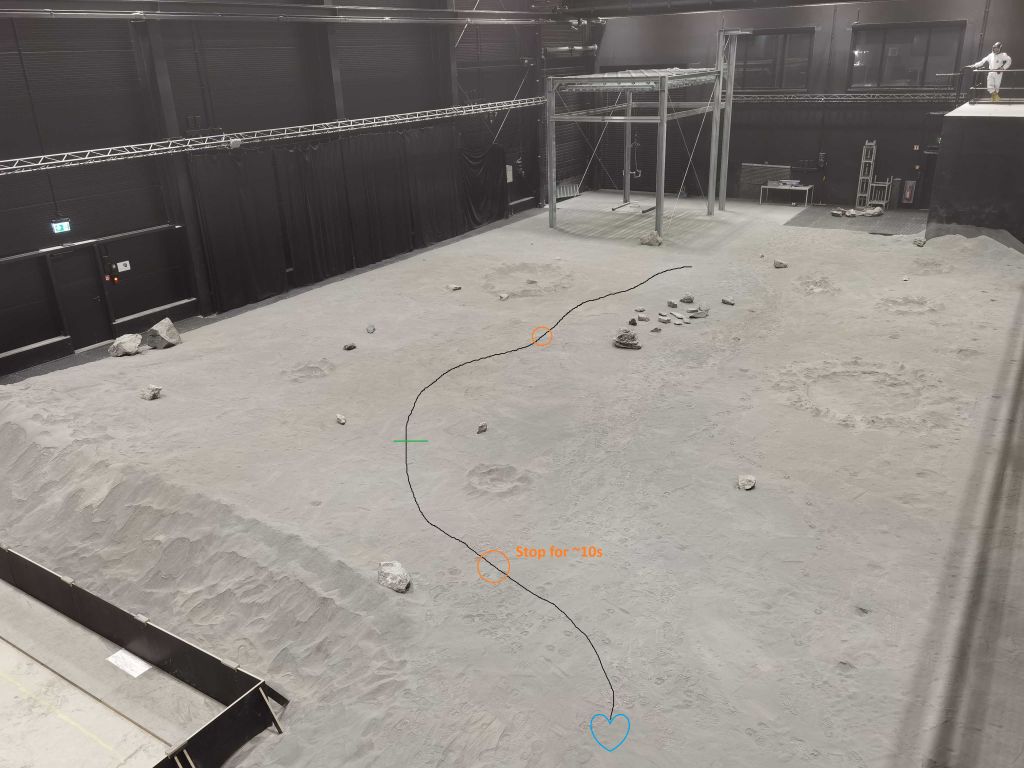}
        \caption{}
    \end{subfigure}\hfill
    \begin{subfigure}[t]{0.24\textwidth}
        \includegraphics[width=\linewidth]{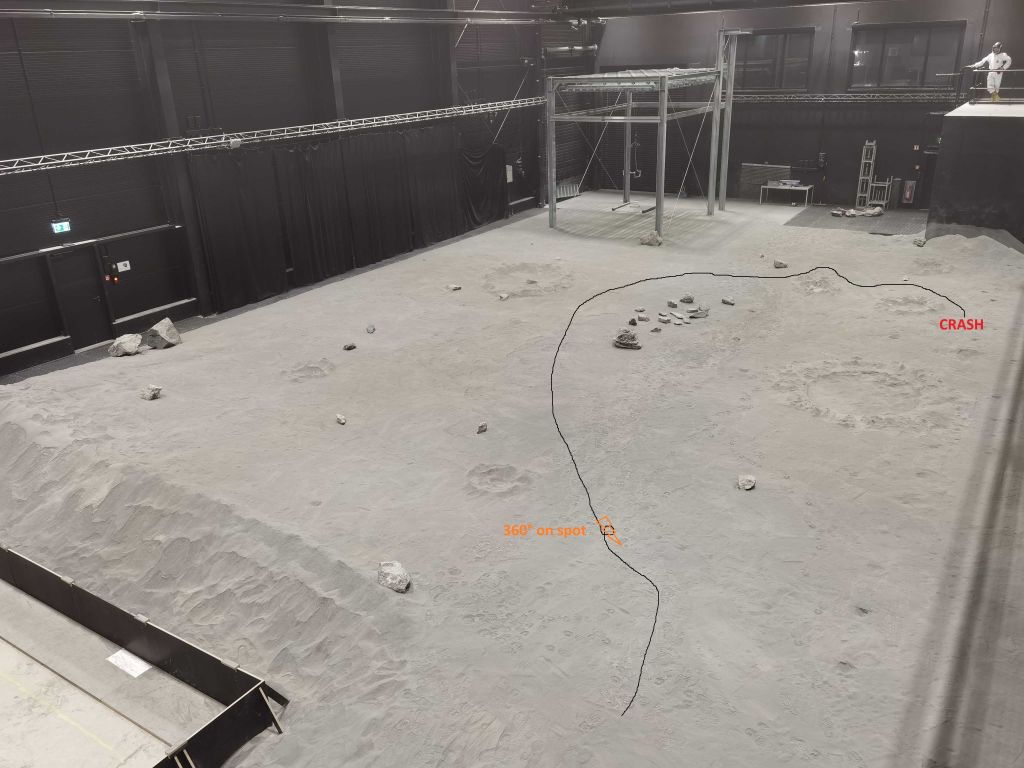}
        \caption{}
    \end{subfigure}

    \vspace{0.5em}

    % Row 2: 3 images
    \begin{subfigure}[t]{0.32\textwidth}
        \includegraphics[width=\linewidth]{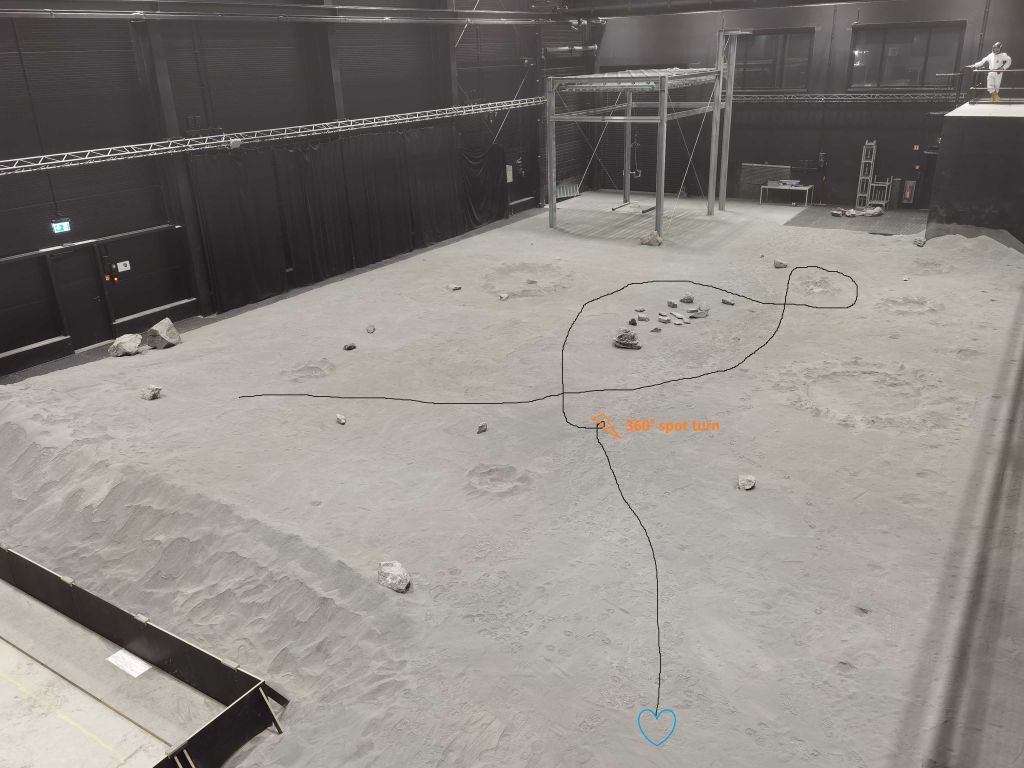}
        \caption{}
    \end{subfigure}\hfill
    \begin{subfigure}[t]{0.32\textwidth}
        \includegraphics[width=\linewidth]{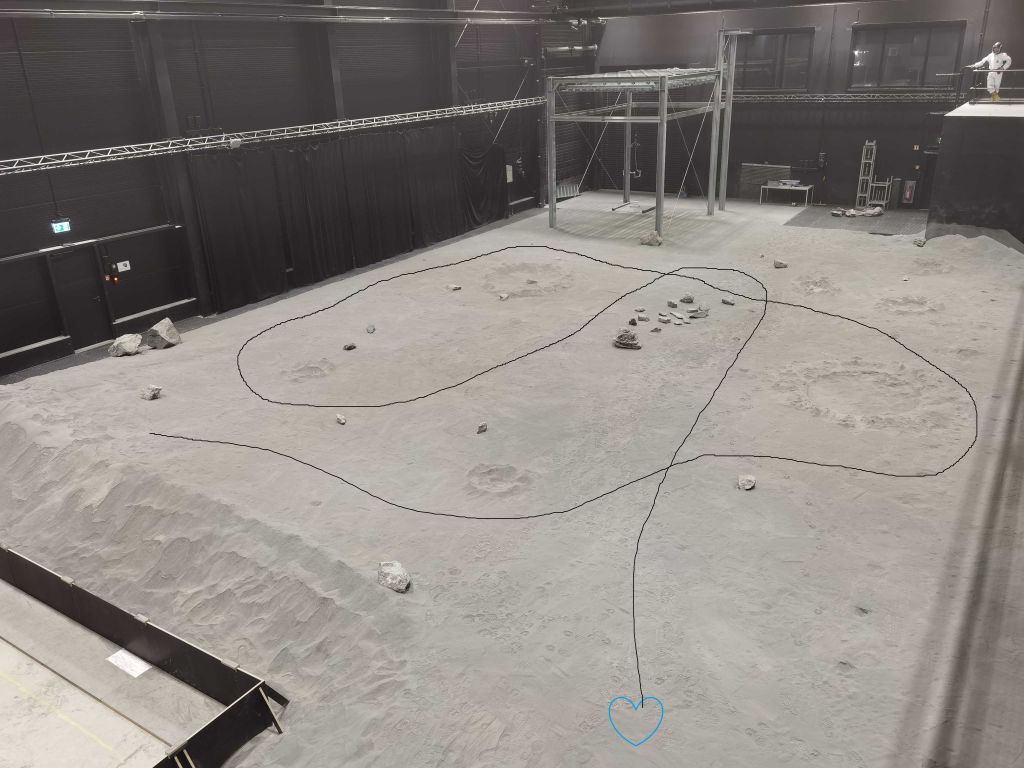}
        \caption{}
    \end{subfigure}\hfill
    \begin{subfigure}[t]{0.32\textwidth}
        \includegraphics[width=\linewidth]{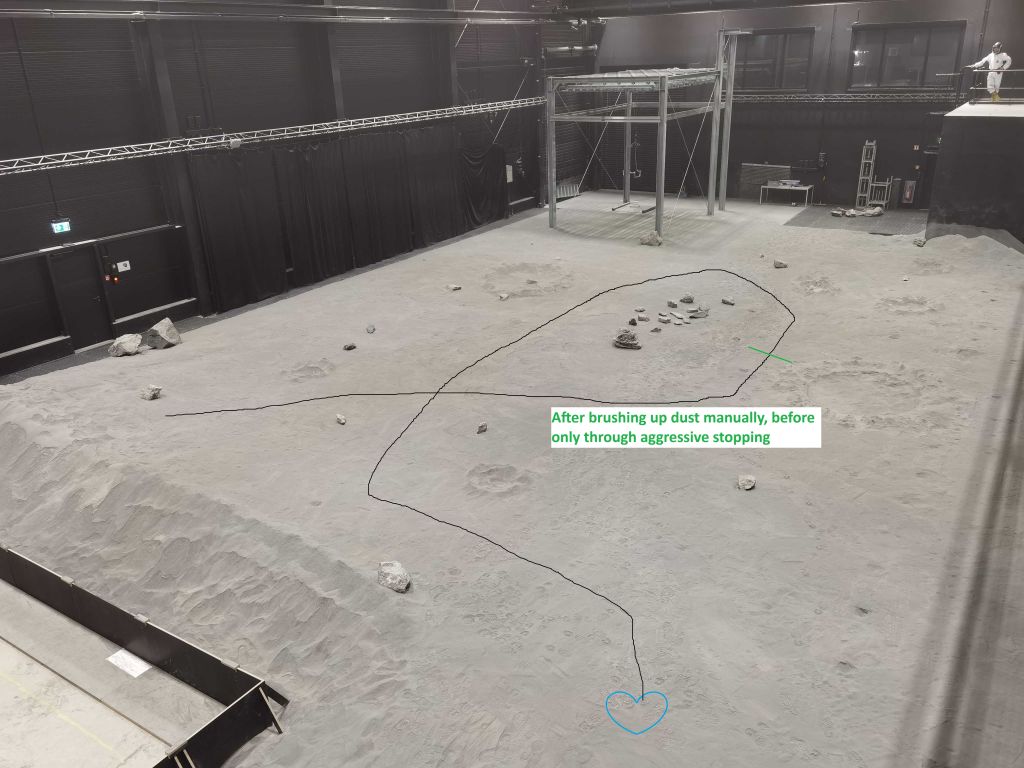}
        \caption{}
    \end{subfigure}

    \vspace{0.5em}

    % Row 3: 4 images
    \begin{subfigure}[t]{0.24\textwidth}
        \includegraphics[width=\linewidth]{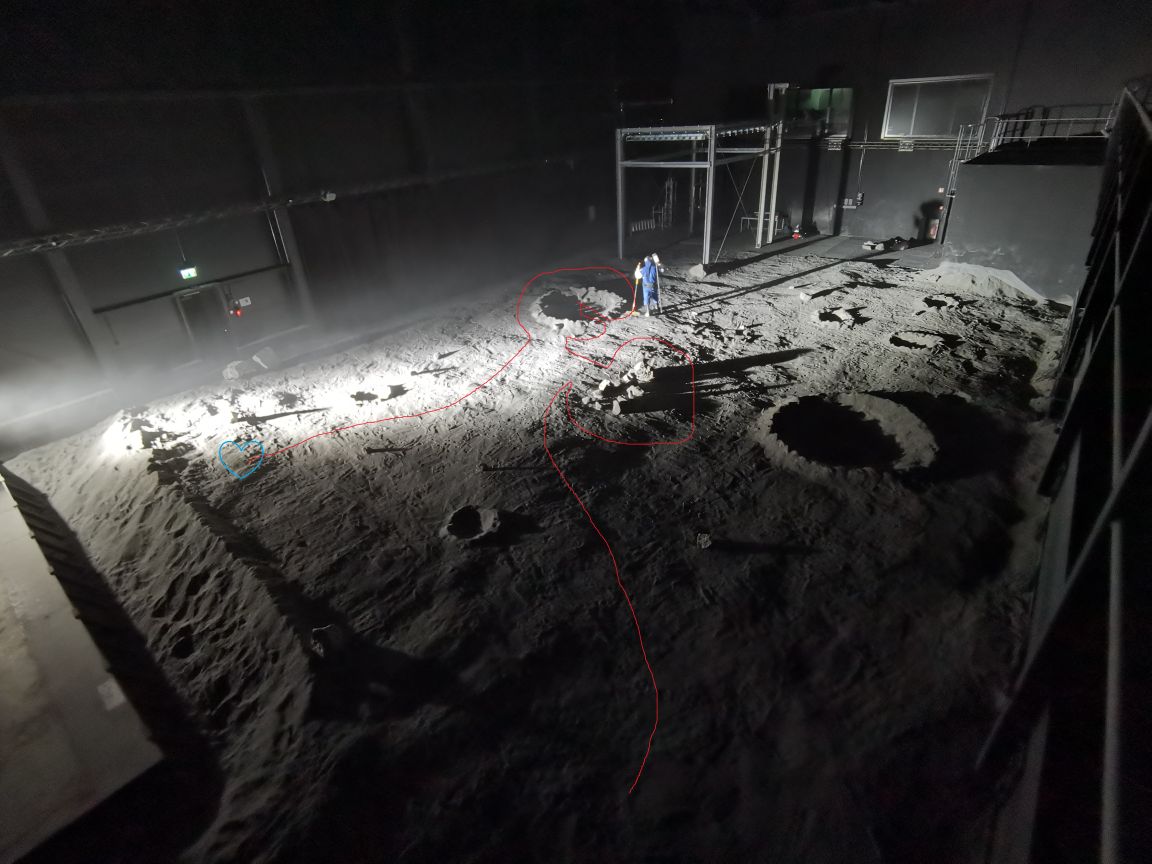}
    \caption{}
    \end{subfigure}\hfill
    \begin{subfigure}[t]{0.24\textwidth}
        \includegraphics[width=\linewidth]{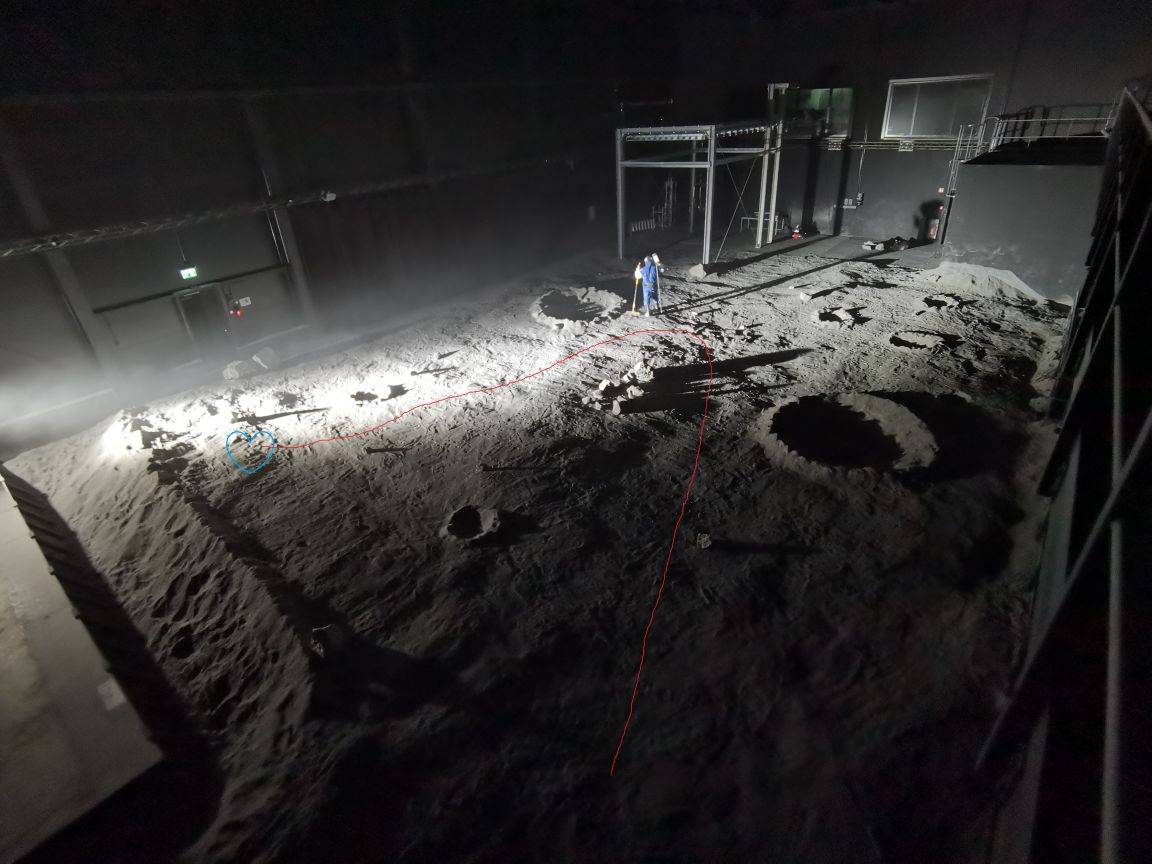}
        \caption{}
    \end{subfigure}\hfill
    \begin{subfigure}[t]{0.24\textwidth}
        \includegraphics[width=\linewidth]{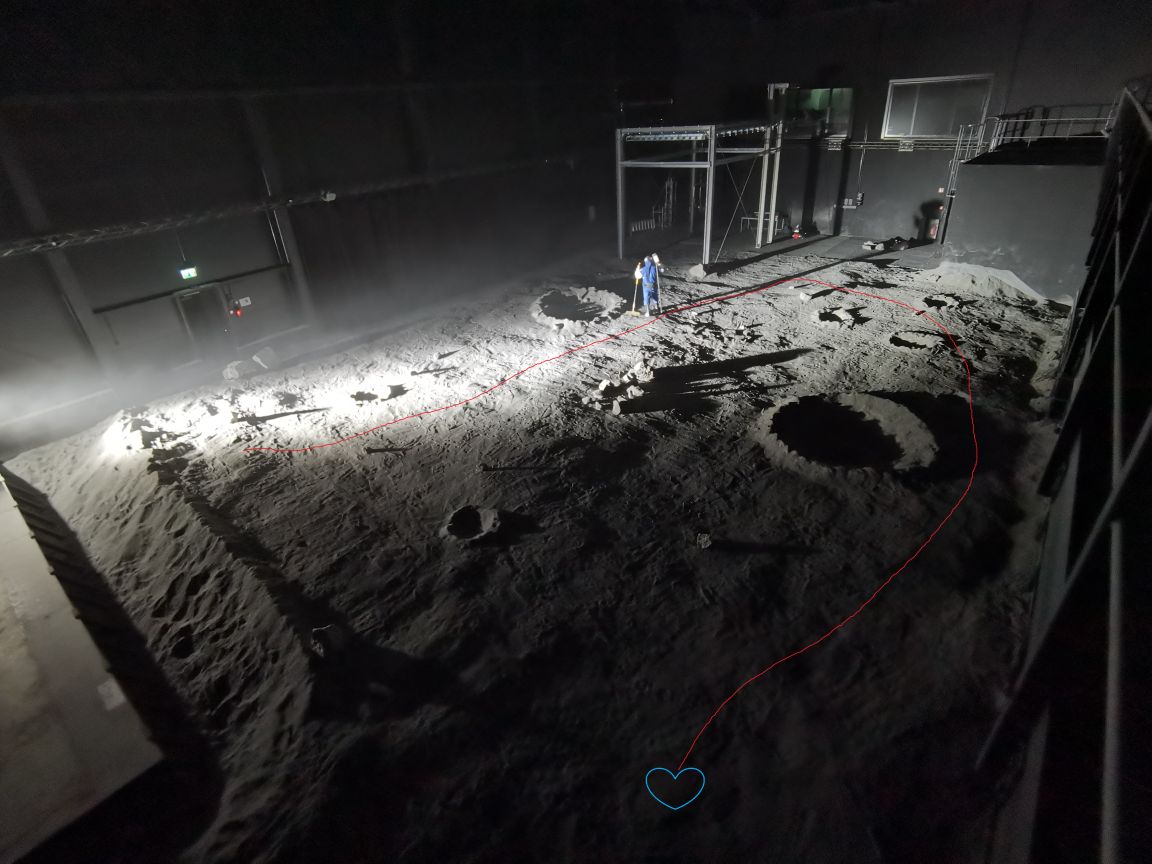}
        \caption{}
    \end{subfigure}\hfill
    \begin{subfigure}[t]{0.24\textwidth}
        \includegraphics[width=\linewidth]{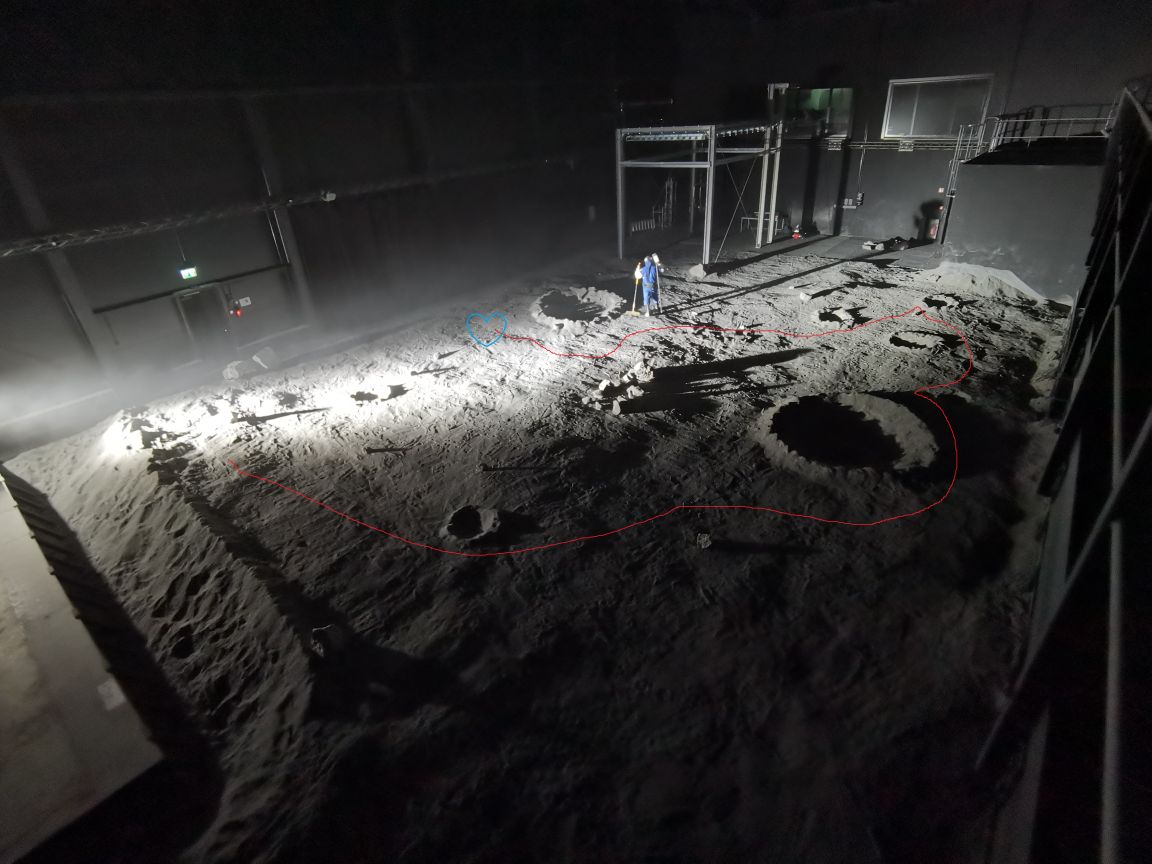}
        \caption{}
    \end{subfigure}

    \caption{Navigation and data collection mission trajectory overviews.}
    \label{fig:anymal-navigation}
\end{figure*}

ANYmal-D with the BOXi was used to evaluate long-horizon locomotion, navigation, and visual-inertial data collection in the LUNA regolith field. 
The experimental evaluation investigates the influence of several camera-related parameters within a lunar analogue environment. In particular, the study focuses on camera resolution, frame rate, shutter type, and the comparison between monochrome and color imaging sensors. These factors are assessed with respect to their impact on visual-inertial navigation performance under representative locomotion and illumination conditions.

\begin{figure*}[!t]
    \centering
    \begin{subfigure}[t]{0.32\textwidth}
        \centering
        \includegraphics[width=\linewidth]{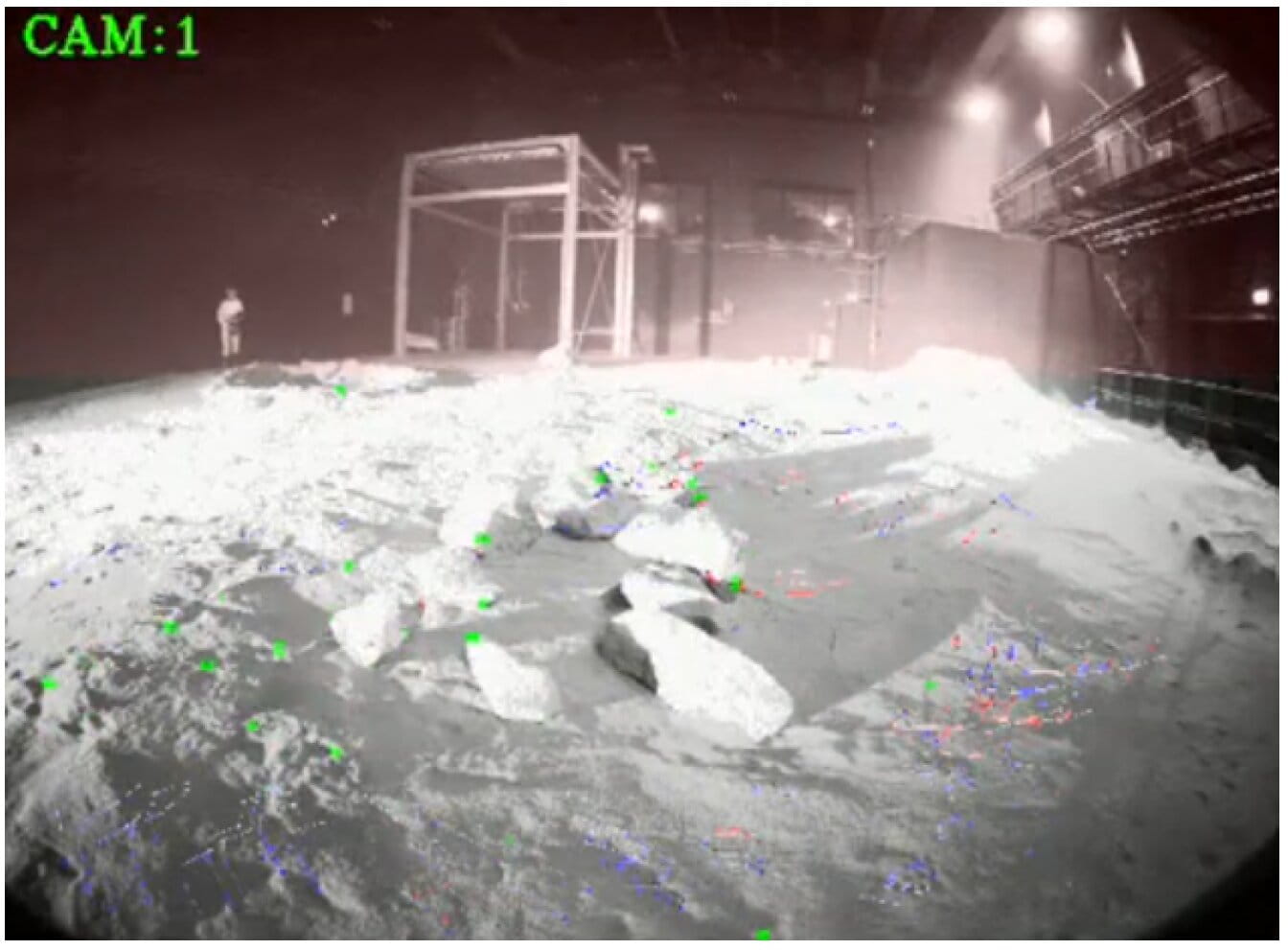}
        \caption{View away from light source.}
        \label{fig:features-away-light}
    \end{subfigure}\hfill
    \begin{subfigure}[t]{0.32\textwidth}
        \centering
        \includegraphics[width=\linewidth]{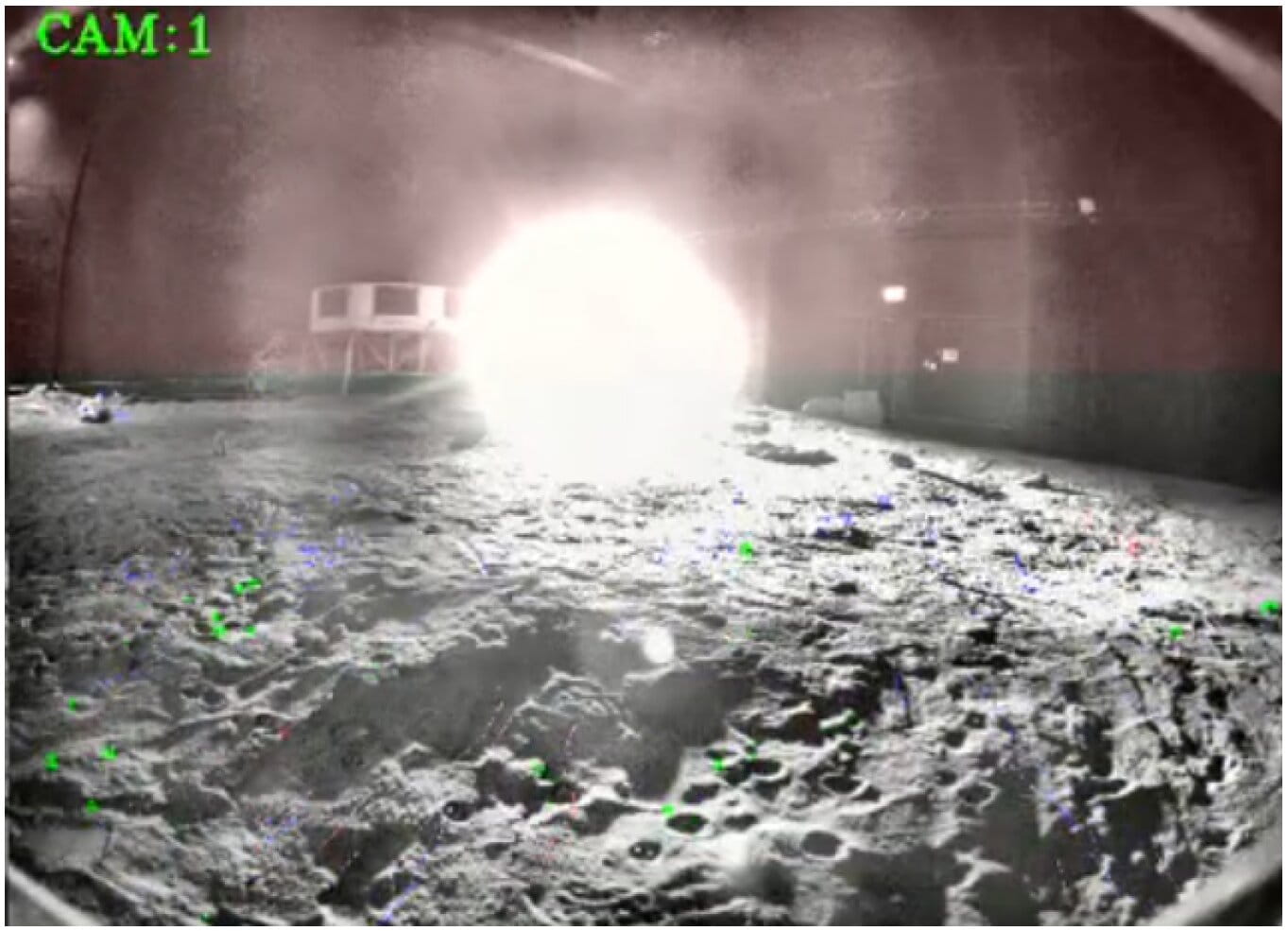}
        \caption{View towards light source.}
        \label{fig:features-to-light}
    \end{subfigure}\hfill
    \begin{subfigure}[t]{0.32\textwidth}
        \centering
        \includegraphics[width=\linewidth]{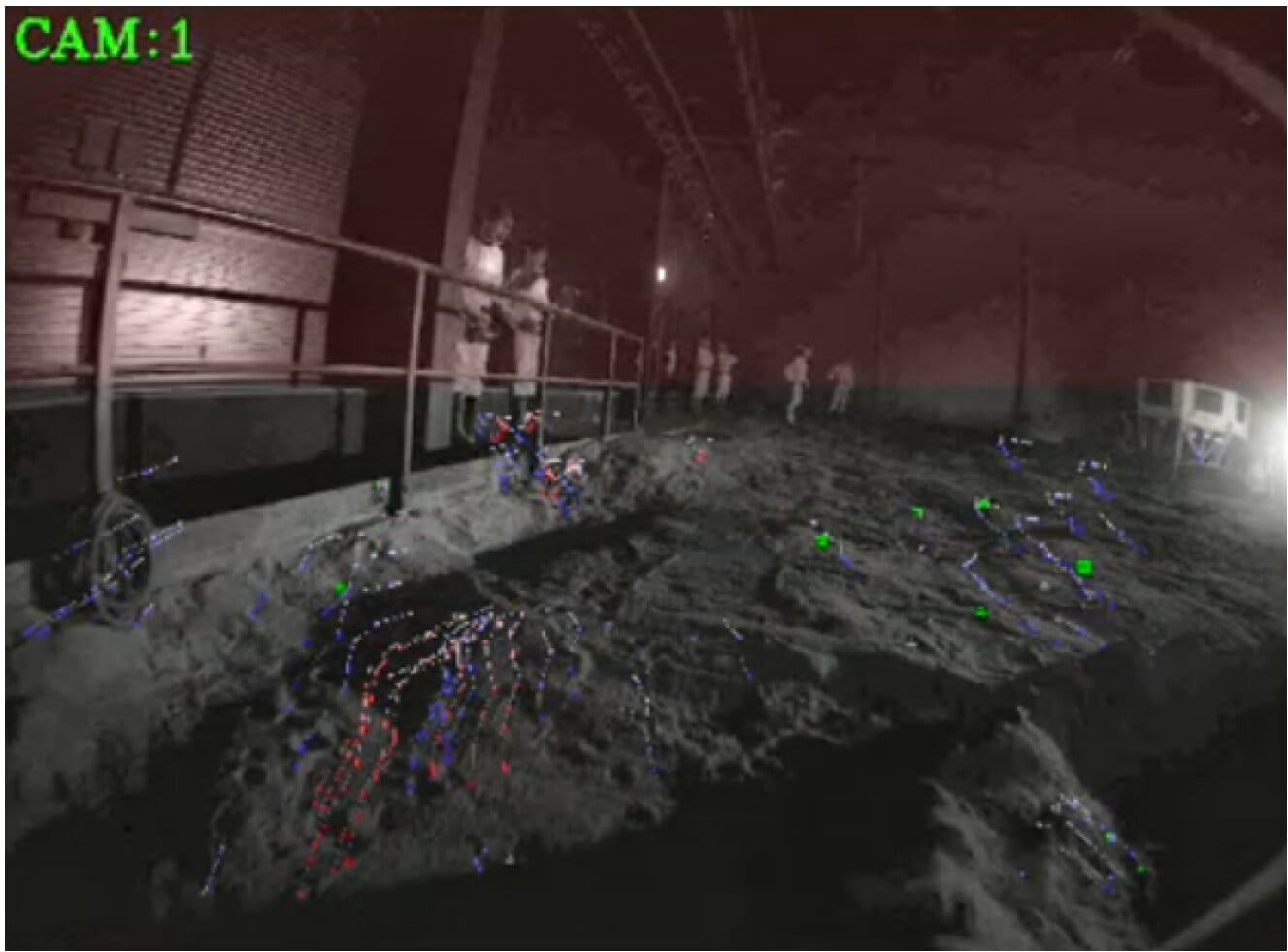}
        \caption{Limited illumination}
        \label{fig:features-shadows}
    \end{subfigure}
    \caption{Tracked features from VIO on stereo monochrome images. Green dots indicate SLAM features, blue dots are tracked in both images, and red dots are tracked in only one image.}
    \label{fig:features-tracked}
\end{figure*}

A total of 11 missions were conducted with different traverse profiles as shown in \cref{fig:anymal-navigation}. The distance walked ranged between \SI{20}{\meter} and \SI{100}{\meter} during the tests. The missions varied in walking pattern (straight, curves, point turns, stops), illumination, terrain (amount and size of boulders and craters, slopes, featureless terrain).
Most features were tracked in the near field, mainly in front of the camera (\cref{fig:features-tracked}). Tracking was generally stable across different lighting conditions, with fewer features in shadowed or darker regions. Performance degraded briefly with dust in the camera view and during rapid motions that introduced motion blur and large inter-frame displacements.

Key limitations included image overexposure when facing the light source, reduced feature quality in shadowed or low-texture regions, dust entering the field of view or accumulating near sensors, and lower frame rates due to onboard compute and real-time processing constraints. These findings also underscore the role of shutter selection and raise the question of whether RGB-only localization is sufficiently reliable under lunar-relevant lighting.
%For this campaign, visual-inertial odometry was computed offline from the recorded datasets due to the limited onboard compute budget. 
These datasets will be useful for future lunar legged robot navigation stacks by supporting studies of illumination robustness, dust tolerance, camera exposure and shutter behavior.

\subsection{Dust, Regolith Disturbance, and Sensor Degradation \label{sec:locomotion-dust}}

Dust generation and regolith disturbance were important qualitative observations during the campaign. Lunar regolith particles are known to be adhesive and abrasive. Therefore, dust accumulation can negatively affect critical surfaces and mechanisms, including solar panels, radiators, optical sensors, joints, seals, and scientific payloads.  The controlled dry regolith bed provided useful system-scale insight, although the EAC-1A simulant in LUNA does not fully reproduce lunar dust behavior since it is in Earth gravity, in an atmosphere, and without electrostatically charged particles.

\begin{figure*}[!t]
    \centering
    \begin{subfigure}[t]{0.32\textwidth}
        \centering
        \includegraphics[width=\linewidth]{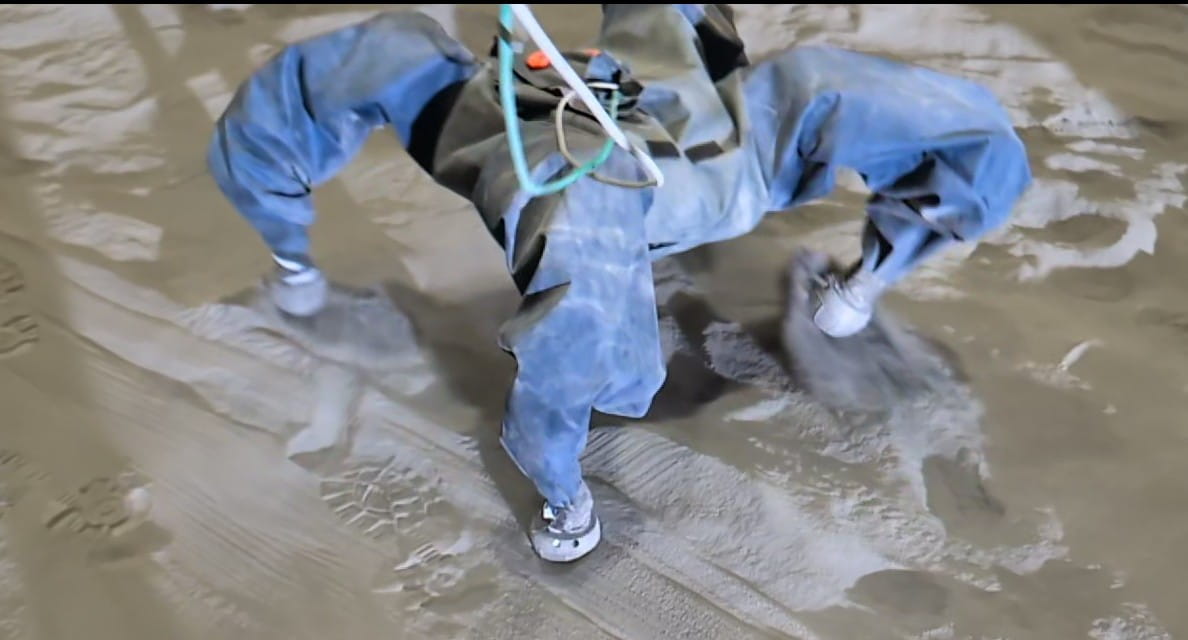}
        \caption{Dust generated during foot drag.}
        \label{fig:dust-drag}
    \end{subfigure}\hfill
    \begin{subfigure}[t]{0.32\textwidth}
        \centering
        \includegraphics[width=\linewidth]{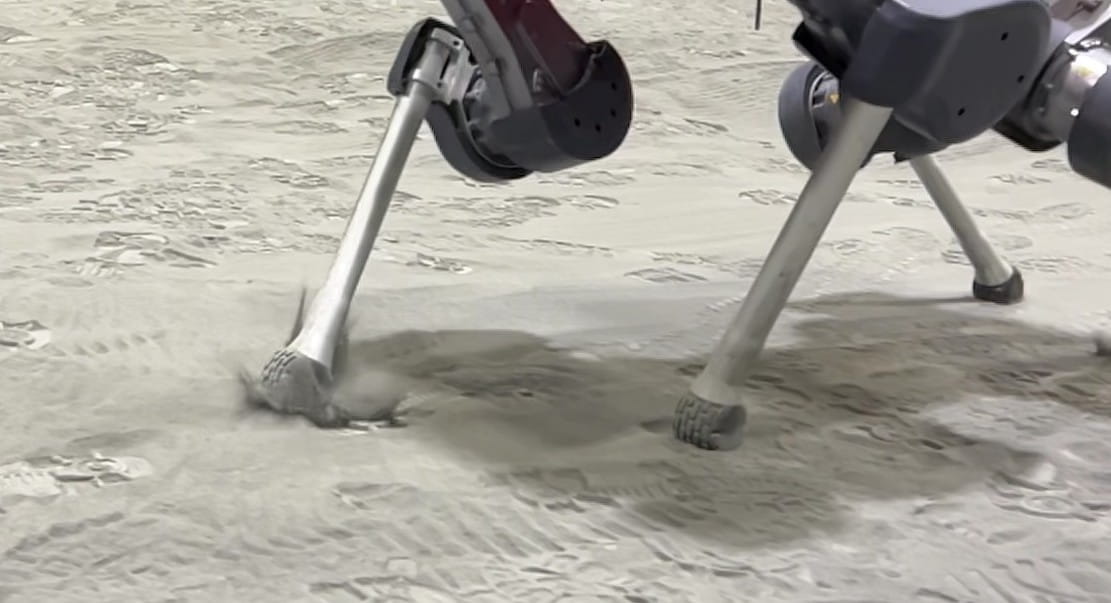}
        \caption{Dust generated during foot impact.}
        \label{fig:dust-stamp}
    \end{subfigure}    \hfill
    \begin{subfigure}[t]{0.32\textwidth}
        \centering
        \includegraphics[width=\linewidth]{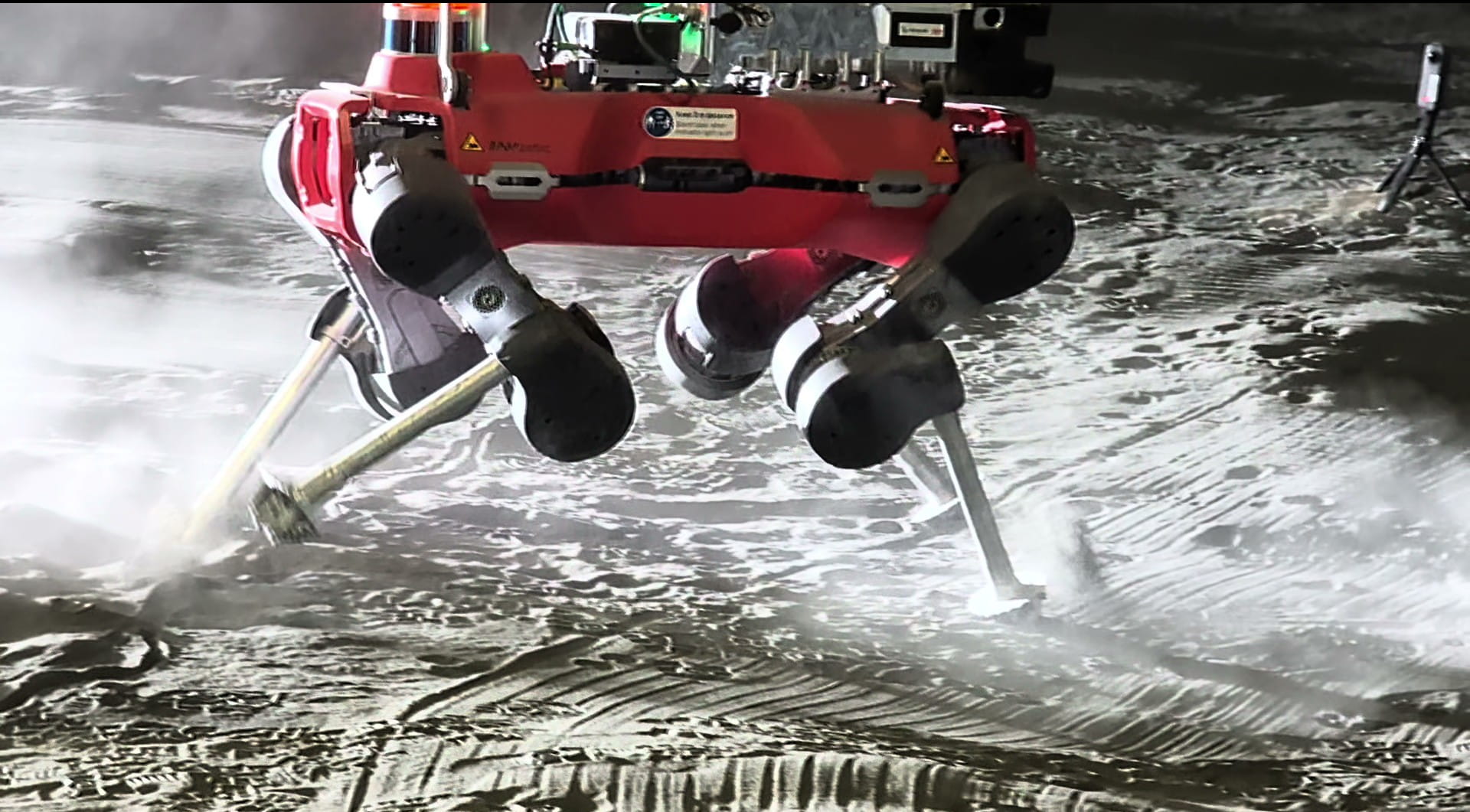}
        \caption{Dust cloud generated by a high-traction foot end-effector.}
        \label{fig:dust-cloud}
    \end{subfigure}

    \caption{Examples of dust generation observed during the LUNA testing campaign.}
    \label{fig:luna-dust}
\end{figure*}

During legged locomotion, dust and simulant displacement were primarily generated through two mechanisms: foot drag and foot impact. Foot drag occurred when the foot moved tangentially across the simulant surface during stance or lift-off, displacing material as shown in \cref{fig:dust-drag}. In some cases, the swing trajectory also caused the particles to be kicked forward as the foot left the ground. This behavior differs from typical wheeled rover motion, in which disturbed material is often ejected rearward by wheel rotation. For legged robots, forward particle ejection increases the self-generated dust exposure.

A simple way to reduce foot-drag-induced dust generation is to use a more vertical foot lift-off trajectory with a greater clearance. However, this may increase swing-leg flight phase motion, locomotion power consumption, and cycle time, potentially limiting walking speed. Dust was also generated during foot impact, as shown in \cref{fig:dust-stamp}, where the kinetic energy of the descending leg was transferred into the simulant at touchdown. These observations suggest that dust-aware locomotion should consider both tangential foot motion and vertical impact velocity, rather than only minimizing slip or maximizing terrain robustness.

The choice of the foot end-effector influenced dust generation. High-traction foot designs with more aggressive contact features tended to scoop and eject larger quantities of regolith simulant, producing visible dust clouds, as shown in \cref{fig:dust-cloud}. This effect was less pronounced with more conventional rounded foot designs, as shown in \cref{fig:dust-drag,fig:dust-stamp}.

Overall, the campaign indicates that future legged lunar robots should treat dust generation as a locomotion-relevant constraint and reward, meaning that future policies and simulations should account for foot-regolith dust interactions. Incorporating these effects into training environments may enable locomotion policies that balance traversability, power efficiency, and dust mitigation for long-duration lunar-surface operations.

\subsection{Operational Constraints and Field Deployment Lessons\label{sec:ops-lessons}}

The LUNA testing campaign revealed several practical constraints that affected test execution, data collection, and repeatability. First, perception systems require an environment-specific masking setup because the indoor facility contains walls and structural elements that are not representative of an open lunar surface. Vision masking was therefore needed during operation or post-processing to prevent mapping and navigation pipelines from using facility-specific features.

Facility access and safety procedures also introduced operational overhead. Entry into the test hall was controlled, only a limited number of personnel were allowed inside, and all personnel were required to wear protective equipment. Communication between personnel inside and outside the hall relied on radios, which required clear role assignments, communication protocols, and training prior to the campaign.

The communication infrastructure was another limitation. WiFi connectivity in the hall was unreliable at times during the campaign, and wired LAN connections were used for some experiments. Future campaigns would benefit from a more distributed wireless network setup around the test field.

Finally, terrain repeatability was limited by manual preparation of the regolith bed. In loose regolith and crater-like regions, local packing conditions could vary between trials, affecting slip, sinkage, and dust generation. Additionally, the regolith should be deposited in a homogeneous way to limit the number of features. Careful terrain conditioning and documentation are therefore necessary for more repeatable locomotion experiments.

\section{Remaining Gaps Toward Lunar Deployment}

\subsection{Regolith- and Dust-Aware Locomotion Policy}

As discussed in \cref{sec:loose-regolith,sec:locomotion-dust}, RL policies for lunar legged robots should account for granular contact dynamics, foot sinkage, drag, and dust generation. 
These effects are not only terrain-interaction issues, but also directly influence locomotion robustness, power consumption, dust exposure, and long-duration operability.

However, incorporating these effects into RL training remains a challenge. Most GPU-accelerated training environments rely on simplified rigid-body contact models, while high-fidelity granular simulations are often too computationally expensive for large-scale policy optimization. Existing approaches for modeling granular contact in legged locomotion provide useful starting points, but gaps remain in simulation fidelity, contact geometry assumptions, and transferability to lunar-relevant regolith conditions \cite{choi2023learning, luo2025mild}.

Domain randomization is also more difficult for granular terrain than for rigid terrain. Granular contact specific parameters are coupled to foot geometry, impact velocity, gravity direction, and terrain preparation \cite{li2013terradynamics}. These interactions are further complicated by reduced gravity, non-quasi-static contacts, and limited terradynamics data. Future work should therefore combine improved granular contact models, targeted terradynamics experiments, and domain randomization strategies that explicitly capture regolith- and dust-related uncertainty.

\subsection{Illumination-Aware Navigation}
As observed in \cref{sec:long-horizon-nav}, lunar-analog illumination can degrade visual localization through overexposure, sharp shadows, low-texture terrain, and changing surface appearance. These effects can be further worsened by dust clouds or accumulated dust on optical surfaces. Future lunar navigation stacks should therefore account for both illumination variation and sensor degradation during operation.

This requires robust exposure control, appropriate shutter selection, high-dynamic-range sensing, and sensor fusion beyond RGB imagery alone. Perception pipelines should also detect unreliable visual conditions, such as shadow boundaries, low feature density, dust in the field of view, or degraded optics, and switch to more conservative localization or motion strategies. Future analog tests should include repeatable lighting and dust-exposure conditions to quantify these failure modes.

\subsection{Mission-Level Navigation and CONOPS}

Mission-level navigation and CONOPS require dedicated verification beyond short-range locomotion tests. Future lunar legged robots may need to operate over sub-kilometer mission scales~\cite{kolvenbach2025lunarleaper}, where performance depends not only on traversability, but also on reliable localization, route execution, operator interaction, and recovery behavior. 

Scientific missions also impose requirements beyond reaching a target location. Robots must be able to collect high-quality and repeatable measurements both during traversal and at points of interest. This may require standing stationary for extended periods, accurately localizing measurement sites, orienting instruments or sensors in the body at desired angles, and minimizing motion, vibration, or actuator noise during data acquisition.

% Therefore, future analog campaigns should evaluate legged robots as mission systems rather than only as locomotion platforms. Relevant metrics include traverse completion, localization drift, operator intervention frequency, time spent in scientific measurement poses, communication robustness, and the ability to recover from navigation or locomotion failures under realistic operational constraints.

\subsection{Toward Higher Technology Readiness}

The legged robots tested in this campaign are currently limited to approximately TRL 4-5, and neither ANYmal-D nor Magnecko was designed for planetary surface environments. Advancing toward higher TRL will require a clean-sheet dedicated lunar robot design that addresses the specific power profile of legged locomotion, including high peak actuator loads, thermal and dust management for exposed leg joints, and other space-specific design constraints.

Regolith-aware RL policies also require higher-fidelity hardware validation. A key challenge is that no single terrestrial facility can fully replicate the lunar environment. Therefore, future validation should combine complementary tests, including thermal-vacuum locomotion without regolith, LUNA campaigns with regolith simulant, vacuum dust-chamber tests, gravity-offloading experiments, and large-scale outdoor field trials for mission-level autonomy and CONOPS.

\section{Conclusion}

% This paper presented lessons learned from the LUNA 2025 campaign, in which ANYmal and Magnecko were deployed at the ESA/DLR LUNA lunar analog facility to evaluate legged robotic systems beyond controlled laboratory conditions. 
% The campaign showed potential that quadrupedal robots can provide useful mobility on regolith-like terrain, including loose soil, crater-like regions, and obstacle areas, but also revealed that lunar readiness is limited by system-level effects that are difficult to expose in simulation alone. In particular, foot-regolith interaction, dust generation, illumination-dependent perception, terrain preparation, communication constraints, and operator procedures all affected deployment performance. These observations indicate that the next step toward lunar legged robots is not only improved locomotion or space-compatible hardware, but tighter integration between regolith- and dust-aware policy learning and illumination-aware perception, repeatable analogue testing, and mission-level operational validation.

This paper presented lessons from the 2025 field deployment of the LUNA testing campaign, in which ANYmal-D and Magnecko were operated at the LUNA facility. The campaign showed that quadrupedal robots can traverse lunar regolith simulant, including loose soil, crater-like terrain, and obstacle-rich areas, but also revealed system-level limitations that are difficult to expose in laboratory tests or rigid-contact simulations alone.

The main lessons are that foot-regolith interaction affects locomotion through sinkage, slip, and dust generation; lunar-analogue illumination can degrade visual localization through overexposure, shadows, and low-texture regions; and field deployment is strongly affected by terrain preparation, communication, masking, safety procedures, and repeatability. These observations suggest that future lunar legged robots should be developed through tighter integration of regolith-aware locomotion RL policy, illumination-robust perception, repeatable analogue testing, and mission-level operational validation.

\section*{ACKNOWLEDGMENTS}

We thank the \textit{Magnecko} team for their continued support with software and hardware maintenance, and, more generally, for building a well-engineered and easy-to-use robotic system. We also thank the ESA/DLR LUNA facility team for providing access to the lunar analogue test field and for their support with facility preparation, safety procedures, terrain setup, and test operations during the campaign.
LunarLeaper acknowledges funding through the CHiwi Foundation, SERI's NASO program (1-012571, LunarLeaper 2024), Moog Inc., maxon, and MARVIS funding.

\bibliographystyle{IEEEtran}
\bibliography{references_lunarjumping, references}
\end{document}